\documentclass[sigconf]{acmart}

\usepackage{url}            
\usepackage{booktabs}       
\usepackage{amsfonts}       
\usepackage{nicefrac}       
\usepackage{microtype}      
\usepackage{bm}
\usepackage{amsmath}
\usepackage{amsthm}
\usepackage{xcolor}
\usepackage{mathtools}
\usepackage{xspace}
\usepackage{graphicx}
\usepackage{siunitx}
\usepackage{subcaption}
\usepackage{float}
\usepackage{amsthm}

\newcommand{\ntransfer}{\ensuremath{n_\text{T}}}
\newcommand{\abs}[1]{\ensuremath{\lvert#1\rvert}}
\renewcommand{\vec}[1]{\ensuremath{\mathbf{#1}}}
\newcommand{\W}{\ensuremath{\mathbf{W}}\xspace}
\newcommand{\fs}{\ensuremath{f^{\text{S}}}\xspace}
\newcommand{\ft}{\ensuremath{f^{\text{T}}}\xspace}
\newcommand{\fhats}{\ensuremath{\hat{f}^{\mkern1mu\text{S}}}\xspace}
\newcommand{\fhatt}{\ensuremath{\hat{f}^{\mkern1mu\text{T}}}\xspace}
\newcommand{\SO}[1]{\ensuremath{\operatorname{SO}\!\left(#1\right)}\xspace}

\newcommand{\R}{\ensuremath{\mathbb{R}}\xspace}
\newcommand{\vx}{\ensuremath{\vec{x}}\xspace}
\renewcommand{\vv}{\ensuremath{\vec{v}}\xspace}

\newtheorem*{remark}{Remark}

\newcommand{\Reviewer}[1]{}

\AtBeginDocument{%
  }

\copyrightyear{2025}
\acmYear{2025}
\setcopyright{rightsretained}
\acmConference[GECCO '25 Companion]{Genetic and Evolutionary Computation Conference}{July 14--18, 2025}{Malaga, Spain}
\acmBooktitle{Genetic and Evolutionary Computation Conference (GECCO '25 Companion), July 14--18, 2025, Malaga, Spain}\acmDOI{10.1145/3712255.3734291}
\acmISBN{979-8-4007-1464-1/2025/07}

\begin{document}

\title{Transfer Learning of Surrogate Models: Integrating Domain Warping and Affine Transformations}

\author{Shuaiqun Pan}
\orcid{0000000170394875}
\affiliation{%
  \institution{LIACS, Leiden University}
  \city{Leiden}
  \country{The Netherlands}}
\email{s.pan@liacs.leidenuniv.nl}

\author{Diederick Vermetten}
\orcid{0000000330407162}
\affiliation{%
  \institution{LIACS, Leiden University}
  \city{Leiden}
  \country{The Netherlands}}
\email{d.l.vermetten@liacs.leidenuniv.nl}

\author{Manuel López-Ibáñez}
\orcid{0000000199741295}
\affiliation{%
  \institution{University of Manchester}
  \city{Manchester}
  \country{UK}}
\email{manuel.lopez-ibanez@manchester.ac.uk}

\author{Thomas B{\"a}ck}
\orcid{0000000167681478}
\affiliation{%
\institution{LIACS, Leiden University}
  \city{Leiden}
  \country{The Netherlands}}
\email{t.h.w.baeck@liacs.leidenuniv.nl}

\author{Hao Wang}
\orcid{0000000249335181}
\affiliation{%
  \institution{LIACS, Leiden University}
  \city{Leiden}
  \country{The Netherlands}}
\email{h.wang@liacs.leidenuniv.nl}


\begin{abstract}
Surrogate models provide efficient alternatives to computationally demanding real-world processes but often require large datasets for effective training. A promising solution to this limitation is the transfer of pre-trained surrogate models to new tasks. Previous studies have investigated the transfer of differentiable and non-differentiable surrogate models, typically assuming an affine transformation between the source and target functions. This paper extends previous research by addressing a broader range of transformations, including linear and nonlinear variations. Specifically, we consider the combination of an unknown input warping—such as one modeled by the beta cumulative distribution function—with an unspecified affine transformation. Our approach achieves transfer learning by employing a limited number of data points from the target task to optimize these transformations, minimizing empirical loss on the transfer dataset. We validate the proposed method on the widely used Black-Box Optimization Benchmark (BBOB) testbed and a real-world transfer learning task from the automobile industry. The results underscore the significant advantages of the approach, revealing that the transferred surrogate significantly outperforms both the original surrogate and the one built from scratch using the transfer dataset, particularly in data-scarce scenarios.
\end{abstract}



\maketitle

\section{Introduction} \label{sec:intro}
Surrogate modeling~\cite{DBLP:books/daglib/0022623, forrester2009recent, DBLP:journals/cce/BhosekarI18, DBLP:journals/isci/TongHMY21} is extensively used to replace expensive simulators for reducing computational costs, for instance in automobile industry~\cite{kiani2016comparative, fang2017design, qiu2018crashworthiness, DBLP:conf/ijcci/ThomaserVBK23}.
Machine learning models are commonly used as surrogates, e.g., Gaussian process regression (GPR)~\cite{satria2020gaussian, rajaram2020deep, DBLP:journals/technometrics/Pourmohamad21} and random forest~\cite{DBLP:journals/tcyb/WangJ20, DBLP:journals/ress/AntoniadisLP21}.
Yet building a new surrogate typically requires many expensive samples, in particular when there are many independent variables. Acquiring this data usually implies running expensive simulations or real-world experiments.
Therefore, we wish to avoid the cost of acquiring large data sets to build surrogates on a new problem instance. Transfer learning~\cite{DBLP:journals/tkde/PanY10, DBLP:journals/jbd/WeissK016, DBLP:journals/pieee/ZhuangQDXZZXH21} can be used to tackle this issue: with a tiny transfer data set sampled on a new problem (the target), we can learn to tweak an accurate surrogate trained on an old problem (the source), provided certain symmetry/invariances between problems.

Covariance shift~\cite{SugiyamaKM07,SHIMODAIRA2000227,PathakMW22} is an important type of symmetry, which says that for a regression task to approximate the source function $\fs\colon\R^d\rightarrow\R$, a target function $\ft$ can be obtained from $\fs$ by transforming the domain thereof. Namely, there exists a bijection $g\colon \R^d \rightarrow \R^d$ such that $\ft = \fs \circ g$. As for the surrogate modeling, the covariance shift implies the predictive distribution $P(y|\vec{x})$ remains unchanged between the source and the target while $P(\vec{x})$ differs. Previous studies~\cite{PanVerLopBac2024transfer, pan2025transferlearningsurrogatemodels} have specialized with $g$ being an affine transformation, estimating the parameters by minimizing a loss function on a tiny transfer data evaluated on the target.

\begin{figure*}[!ht]
\centering
\begin{tabular}{*{7}{>{\centering\arraybackslash\small}p{.118\textwidth}}}
Source function & Target function & Original GPR & Trained with\newline 40 samples & Transferred with 40 samples & Trained with\newline 80 samples & Transferred with 80 samples \\
\end{tabular}
\includegraphics[width=\textwidth]{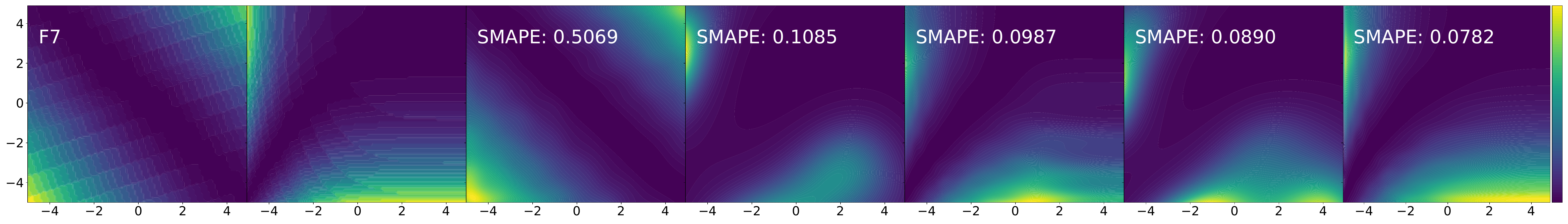}
\caption{For the 2D F7 StepEllipsoid function, we show, from left to right: the contour lines of the source function ($\fs$), the target (\ft), the original GPR (\fhats) trained to approximate \fs, the GPR model trained from scratch with 40 samples from \ft, and the transferred GPR using the same samples. \ft is created from \fs by transforming the domain thereof with affine warping. We show the transfer effect with 80 data points in the last two subplots.
}
\Description{Illustration of the impact of our transfer learning method applied to a Gaussian process regression (GPR) model on a 2D BBOB function, using F7 StepEllipsoid as an example.}
\label{figure:transferGPR_BBOB}
\end{figure*}

However, the affine transformation $g$ might be too restrictive to model complex real-world scenarios, e.g., non-linearity is necessary in the automobile industry problems~\cite{pan2025transferlearningsurrogatemodels}. Hence, we propose implementing a non-linear function $g$ with a beta cumulative distribution function (CDF)~\cite{pmlr-v32-snoek14}. We showcase the preliminary results of our method in Figure~\ref{figure:transferGPR_BBOB}, where we train a GPR on function F7 from the BBOB benchmark suite. Comparing the transferred model to the one trained from scratch on the target function (fourth and fifth subplots), we see the transferred model is more accurate and resembles the contour lines of the target function.   
Our contributions are:
\begin{itemize}
    \item We introduce the non-linear input warping (beta CDF) to affine transfer learning proposed in~\cite{PanVerLopBac2024transfer}, realizing a novel non-linear domain transfer method. 
    \item We numerically validate the effectiveness of our approach on synthetic transfer learning problems created from the Black-Box Optimization Benchmark (BBOB) suite and on a real-world application in the automobile industry.
    \item We analyze the benchmarking outcomes to identify the conditions under which the transferred model outperforms the one trained from scratch.
\end{itemize}

\section{Related Works}\label{sec:related-work}
\paragraph{Transfer learning for GPR}
\citet{saida2023transfer} investigated GPR surrogate model transfer for structural reliability under uncertainties by augmenting the feature space~\cite{DBLP:journals/corr/abs-0907-1815}.
\citet{DBLP:journals/ijon/ZhangYJQ23} proposed a novel strategy that employs a geodesic flow kernel and knee point-based manifold learning to refine GPRs using high-quality knee solutions from previous tasks, thereby enriching training data and boosting solution precision. 
In multi-task settings,
\citet{DBLP:conf/aaai/CaoPZYY10} developed the Adaptive Transfer Learning algorithm (AT-GP) using a semi-parametric transfer kernel. The Transfer Bayesian Committee Machine (Tr-BCM)~\cite{DBLP:journals/kbs/DaOGFL19} introduced a scalable transfer learning approach by aggregating predictions from lightweight local experts, relaxing assumptions of uniform similarity between tasks. Recent advances include \citet{DBLP:journals/kbs/PapezQ22}, which proposed a probabilistic predictor for global source-target interactions, and \citet{DBLP:journals/pami/WeiVQOM23}, which developed an interpretable multi-source transfer kernel for improved cross-task performance.

\paragraph{Transfer learning with input warping} 
\citet{pmlr-v32-snoek14} proposed warping the independent variables with beta CDF to realize non-stationary kernels, where the shape parameters of the beta distribution are inferred with Bayesian estimation (using log-normal priors). Their posterior predictive distribution is obtained by marginalizing the shape parameters. In contrast, we use the beta CDF to capture the non-linear mapping between source and target domains, fitting its shape parameters directly through loss minimization. \citet{cowen2022hebo} introduced Kumaraswamy input warping as a fast, flexible stand‐in for the beta CDF. \citet{DBLP:conf/nips/DuKSP17} presented a Hypothesis Transfer Learning framework linking domains via transformation functions, and \citet{DBLP:journals/tetc/ZhuWPL23} proposed a nonlinear, data‐agnostic mapping to align marginal distributions.

\section{Learning Affine Warping to transfer domains}\label{sec:method}
\subsubsection*{Context} We consider a source regression task: a source function $\fs\colon \mathbb{R}^d\to \mathbb{R}$ to generate the regression data and a trained surrogate model $\fhats$ that approximates $\fs$ accurately. Consider a new regression task $\ft$, the target task. We assume that there exists an unknown nonlinear symmetry between $\fs$ and $\ft$: $\forall \vx\in\R^d, \ft(\vx) = \fs \circ g(\vx)$, $g(\vx) = \mathbf{W}\phi(\vx) + \vv$,  where  $\vv\in\R^d$, $\mathbf{W} \in \SO{d}$ (the rotation group of dimension $d$), and $\phi\colon \R^d \to\R^d$ is a non-linear diffeomorphism. 

\subsubsection*{Goal} We wish to transfer the source surrogate model $\fhats$ to the target function $\ft$ by re-parameterizing it as $\fhats(\mathbf{W}\phi(\vx) + \vv)$ and learn the unknown parameters $\mathbf{W}$, $\vv$, and $\phi$ with a small transfer data set $\mathcal{T} = \{(\mathbf{x}^k, \ft(\mathbf{x}^k))\}_{k=1}^{\ntransfer}$ from the target function.

\subsubsection*{Method} 
\begin{figure}[!htbp]
  \centering
  \includegraphics[width=0.45\textwidth]{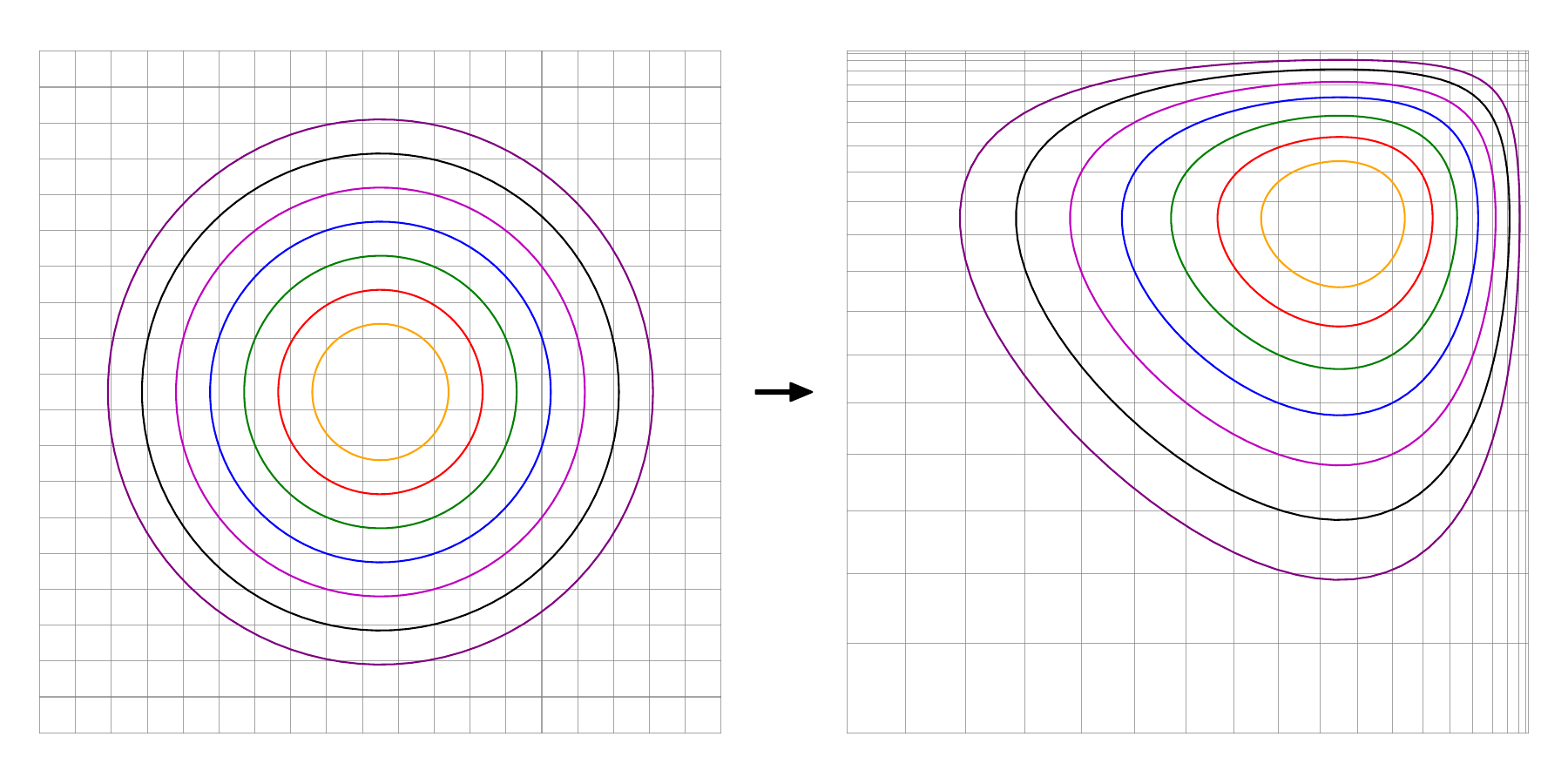}
  \caption{2D input warping: The coordinate system is transformed from left to right by beta CDFs with shape parameter $\alpha=1.0558, \beta=1.9339$ for $x$-axis and $\alpha=0.8655,\beta=1.8148$ for $y$-axis. We show the contour lines of a sphere function on the left and its warped version on the right.}
  \Description{This figure demonstrates the transition from an evenly spaced grid within the $[0, 1] \times [0, 1]$ interval (left) to a warped grid (right) achieved via beta CDF.}
  \label{figure:betacdf_exponential_2d}
\end{figure}

For the non-linear function $\phi$, we consider representing it with the beta cumulative distribution function (CDF)~\cite{pmlr-v32-snoek14}:
\begin{align}
&\!\!\phi(\vx; \theta) = (\phi_1(x_1, \alpha_1, \beta_1), \ldots, \phi_i(x_i, \alpha_i, \beta_i), \ldots, \phi_d(x_d, \alpha_d, \beta_d)) \\
&\phi_i(x_i,\alpha_i, \beta_i) = \int_0^{x_i} \frac{u^{\alpha_i-1} (1-u)^{\beta_i-1}}{B(\alpha_i, \beta_i)} \, du \label{eq:beta-cdf}\\
&\theta = (\alpha_1, \beta_1, \ldots, \alpha_i, \beta_i, \ldots \alpha_d, \beta_d)\in\R^{2d}_{>0},
\end{align}
where \( B(\alpha_i, \beta_i) \) is the beta function. The beta CDF parameterization for each dimension has two unknown parameters $\alpha_i>0, \beta_i>0$. 
Figure~\ref{figure:betacdf_exponential_2d} illustrates input warping on a 2D sphere where each axis is warped nonlinearly using its own shape parameters.

\begin{remark}
    (1) The beta CDF preserves the convexity of the surrogate since it is a smooth, monotonic transformation. (2) For non-universal models, e.g., GPR with a fixed kernel, the beta CDF materializes a non-stationary auto-correlation function, increasing such models' expressivity. (3) For universal models, e.g., deep neural networks, the beta CDF can reshape the input space and make some target functions easier to learn, which is similar to the effect of the normalizing flow~\cite{RezendeM15}. 
\end{remark}

On the transfer data set $\mathcal{T}$, we measure the mean squared error of the re-parameterized source model: 
\begin{align}\label{eq:the-problem}
&\mathcal{L} \colon \mathbb{R}^d \thinspace \times \thinspace \SO{d} \thinspace \times \thinspace \mathbb{R}^{2d}_{>0} \to \mathbb{R}\enspace, \\ 
&(\mathbf{v}, \W, \theta) \mapsto \frac{1}{\ntransfer}\sum_{\mathbf{x}\in \mathcal{T}}\left(\fhats\left(\W(\phi(\mathbf{x};\theta)) + \mathbf{v}\right) - \ft(\mathbf{x})\right)^2\enspace. \nonumber
\end{align}
We solve the transfer learning task by minimizing the above MSE loss.
We shall discuss tackling this minimization problem for differentiable and non-differentiable surrogate models.


\subsection{Transfer differentiable surrogates} If the surrogate model is continuously differentiable, e.g., Gaussian process regression or support vector machine, we can minimize Eq.~\eqref{eq:the-problem} with mini-batch gradient descent.
Let $\vec{y}^k=\mathbf{W}\phi(\vec{x}^k;\theta)$ and $y_i^k$ be $i$-th component of $\vec{y}^k$. The gradient of $\mathcal{L}$ is:

\begin{align}
\frac{\partial\mathcal{L}}{\partial v_i} =& \frac{2}{\ntransfer}\sum_{k=1}^{\ntransfer}\left(\fhats(\vec{y}^{k}) - \ft(\mathbf{x}^{k})\right) \frac{\partial \fhats}{\partial y_i^k} \\
\frac{\partial\mathcal{L}}{\partial W_{ij}} =& \frac{2}{\ntransfer}\sum_{k=1}^{\ntransfer}\left(\fhats(\vec{y}^{k}) - \ft(\mathbf{x}^{k})\right) \phi_j(x^k_j, \alpha_j, \beta_j) \frac{\partial \fhats}{\partial y_i^{k}} \\
\frac{\partial \mathcal{L}}{\partial \alpha_i} =& \frac{2}{\ntransfer} \sum_{k=1}^{\ntransfer} \sum_{\ell=1}^d\left( \fhats(\vec{y}^k) - \ft(\mathbf{x}^k) \right)\frac{\partial \fhats}{\partial y_\ell^k} W_{\ell i} \frac{\partial \phi_i(x^k_i, \alpha_i, \beta_i)}{\partial \alpha_i} \\
\frac{\partial \mathcal{L}}{\partial \beta_i} =& \frac{2}{\ntransfer} \sum_{k=1}^{\ntransfer} \sum_{\ell=1}^d\left( \fhats(\vec{y}^k) - \ft(\mathbf{x}^k) \right)\frac{\partial \fhats}{\partial y_\ell^k} W_{\ell i} \frac{\partial \phi_i(x^k_i, \alpha_i, \beta_i)}{\partial \beta_i}
\end{align}
The derivatives $\phi_i$ w.r.t. $\alpha_i$ and $\beta_i$ are
\begin{align}
    \frac{\partial \phi(x_i, \alpha_i, \beta_i)}{\partial \alpha_i} &= A(x_i, \alpha_i,\beta_i) - \phi(x_i; \alpha_i, \beta_i) \frac{\partial}{\partial \alpha_i} \log B(\alpha_i, \beta_i) \\
    \frac{\partial \phi(x_i, \alpha_i, \beta_i)}{\partial \beta_i} &= B(x_i, \alpha_i, \beta_i) - \phi(x_i; \alpha_i, \beta_i) \frac{\partial}{\partial \beta_i} \log B(\alpha_i, \beta_i) \\
    A(x_i, \alpha_i,\beta_i) &= \int_0^{x_i} \frac{\log(u) u^{\alpha_i-1} (1-u)^{\beta_i-1}}{B(\alpha_i, \beta_i)} du \\
    B(x_i, \alpha_i,\beta_i) &= \int_0^{x_i} \frac{\log(1-u) u^{\alpha_i-1} (1-u)^{\beta_i-1}}{B(\alpha_i, \beta_i)} du \\
    \frac{\partial \log B(\alpha_i, \beta_i) }{\partial \alpha_i} &= \psi(\alpha_i) - \psi(\alpha_i + \beta_i) \\
    \frac{\partial \log B(\alpha_i, \beta_i) }{\partial \beta_i} &= \psi(\beta_i) - \psi(\alpha_i + \beta_i)
\end{align}
where \( \psi \) is the digamma function and $\partial\fhats/\partial y_i^k$ can be computed analytically from surrogate's predictor. All the above derivatives live in Euclidean spaces, to which the vanilla gradient descent algorithm can be applied. However, $\W\in\SO{d}$ is a rotation matrix, and it will not remain in $\SO{d}$ if we perform a descent step with Euclidean gradient $\partial\mathcal{L}/\partial W_{ij}$. Hence, we decide to take a Riemannian gradient descent method, which first computes the Riemannian gradient - an orthogonal projection of $\partial\mathcal{L}/\partial W_{ij}$ onto the tangent space of $\SO{d}$ at $\mathbf{W}$~\cite{PanVerLopBac2024transfer}:
\begin{align}
    \nabla_R\mathcal{L}(\mathbf{W}) = \operatorname{P}\left(\frac{\partial\mathcal{L}}{\partial \mathbf{W}}\right), \quad \operatorname{P}(\mathbf{M}) = \mathbf{W}\frac{\mathbf{W}^\top\mathbf{M} - \mathbf{M}^\top\mathbf{W}}{2}\enspace,\label{eq:projection}
\end{align}
Next, a gradient step (geodesic with initial velocity $\nabla_R\mathcal{L}(\mathbf{W})$) on  $\SO{d}$ from $\mathbf{W}$ can be computed by the exponential map:
\begin{equation}\label{eq:exponential-map}
   \operatorname{Exp}_{\mathbf{W}}(\sigma\nabla_R\mathcal{L}(\mathbf{W})) = \mathbf{W}\operatorname{Exp}\left(\sigma\mathbf{W}^\top\nabla_R\mathcal{L}(\mathbf{W})\right) \in \SO{d}, 
\end{equation}
where $\sigma$ is the step-size and $\operatorname{Exp}$ is the matrix exponential. 

\subsection{Transfer non-differentiable surrogates}
We also extend our methodology to non-differentiable models like random forests by tuning parameters with Covariance matrix adaptation evolution strategy (CMA-ES)~\cite{DBLP:reference/sp/EmmerichS018,DBLP:books/sp/06/Hansen06, DBLP:journals/corr/Hansen16a}. CMA-ES can be applied directly to the search space of the translation parameter $\mathbf{v}$ and the beta CDF parameters $\mathbf{\alpha}$ and $\mathbf{\beta}$, which are Euclidean. However, special treatment is needed for \W, which lives in a smooth manifold \SO{d}. To solve this issue, we consider the Lie group representation $\mathfrak{so}(d)=\{\mathbf{A}\in\mathbb{R}^{d\times d}\colon \mathbf{A}^\top = -\mathbf{A}\}$, which is a flat space (with dimension $d(d-1)/2$), and optimize this representation with CMA-ES. A rotation matrix \W can be recovered from its representation $\mathbf{A}$ with the exponential map, i.e., $\W = \operatorname{Exp}(\mathbf{A})$. For each search point $\mathbf{z}\in\mathbb{R}^{d(d-1)/2}$, we have to transform it into a $d\times d$ antisymmetric matrix to preserve the structure of $\mathfrak{so}(d)$: the components in $\mathbf{z}$ are sequentially assigned to the upper triangular of $\mathbf{A}$ (diagonal entries are zero) row by row. The negative value of the transposition of the upper triangular then fills the lower triangular entries.
\begin{figure*}[!htbp]
  \centering
  \includegraphics[width=\textwidth,trim=0 32mm 0 0,clip]{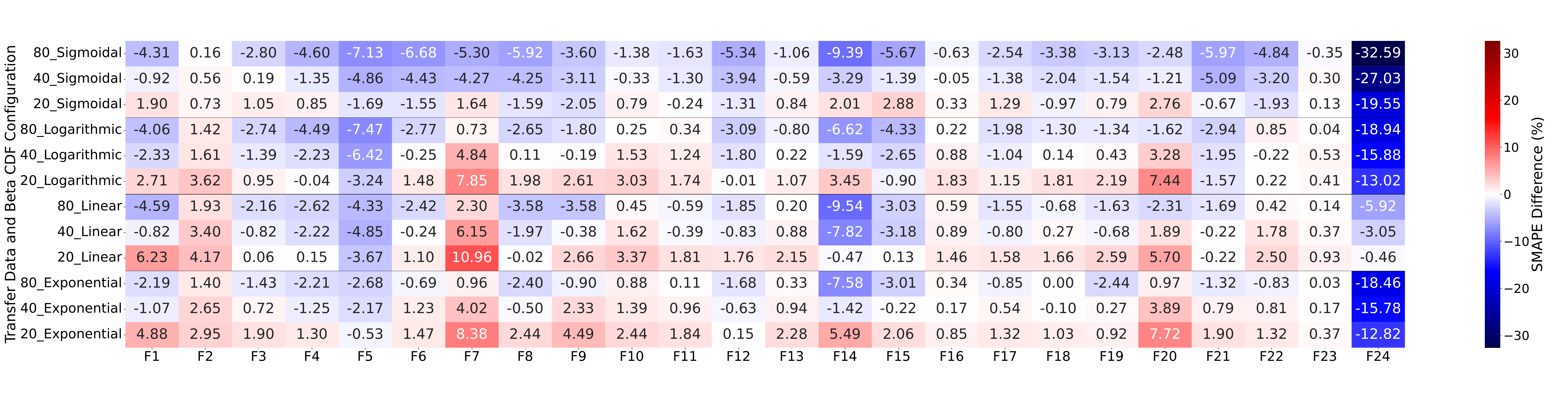}
  \caption{
  On 2D BBOB functions, we compare transferred GPR models with those trained from scratch on the transfer dataset. Each cell displays the percentage difference in average SMAPE (\%) for a combination of BBOB functions, sample size, and beta CDF shape. Positive values (shown in red) indicate superior performance of the transferred model.
  }
  \label{figure:smape_heatmap_2d}
  \Description{Comparison of Gaussian process regression models obtained through transfer learning with those trained from scratch on the transfer dataset, evaluated on 2-dimensional BBOB functions.}
\end{figure*}

\section{Experimental Settings}\label{sec:experiment}

\paragraph{Synthetic tasks based on BBOB}
We first evaluate our method on the Black-Box Optimization Benchmarking (BBOB)~\cite{HanFinRosAug2009bbob,HanAugMer2020coco,bbobfunctions} suite, which consists of 24 continuous, single-objective problems. The BBOB suite has been widely used as a regression benchmark~\cite{singh2018integration, DBLP:journals/tai/TianPZRTJ20, DBLP:conf/nips/ChenSLW0DKKDRPF22, DBLP:conf/emo/YangA23} as it reflects real-world regression difficulties. To create synthetic transfer learning problems out of BBOB, we use the first instance of each BBOB function as the source $\fs$, and construct the target $\ft$ by applying a beta CDF transformation, followed by random rotation and translation transformations, to the base function (so we do not use BBOB's internal instance transformation methods). The shape parameters are sampled from log-normal distributions, i.e., $\log\alpha_i\sim \mathcal{N}(\mu_i^\alpha, \sigma^\alpha_i), \log\beta_i\sim \mathcal{N}(\mu_i^\beta, \sigma_i^\beta)$. Inspired by~\cite{pmlr-v32-snoek14}, we vary the priors to induce different warp geometries: 
\begin{itemize}
    \item linear shape: $\mu_i^\alpha = \mu_i^\beta = 0, \sigma^\alpha_i = \sigma_i^\beta = 0.5$
    \item exponential shape: $\mu_i^\alpha=0, \sigma^\alpha_i=0.25, \mu_i^\beta=1, \sigma_i^\beta=1$
    \item logarithmic shape: $\mu_i^\alpha=1, \sigma^\alpha_i=1, \mu_i^\beta=0, \sigma_i^\beta=0.25$
    \item Sigmoidal shape: $\mu_i^\alpha = \mu_i^\beta = 2, \sigma^\alpha_i = \sigma_i^\beta = 0.5$ 
\end{itemize}






To generate the training data set for $\fhats$, we sample $1\,000 \times d$ points uniformly at random in the domain $[-5,5]^d$ and evaluate them on $\fs$. To assess the performance of the original GPR model $\fhats$ on the target function $\ft$, we create an independent test dataset of the same size, $1\,000 \times d$, sampled uniformly at random from $\ft$. The transfer learning process uses a transfer dataset $\mathcal{T}$, containing $40 \times d$ randomly sampled points from $\ft$. In addition, we consider a minimal transfer dataset of $40$ points, independent of the dimensionality of the target problem, smaller datasets of $20$ points for 2-dimensional and 5-dimensional cases, and a larger dataset of $80$ points for the 10-dimensional problems. After transfer learning, the effectiveness of the transferred GPR model $\fhatt$ is evaluated using the same test set employed for the original GPR model. We also train a GPR model from scratch directly on $\mathcal{T}$ for comparison.

\paragraph{Real-world benchmark from automobile industry}
We validate our transfer approach on an ABS‐braking dataset from automotive engineering~\cite{DBLP:conf/ijcci/ThomaserVBK23}. The search space consists of two ABS parameters, $x_1$ and $x_2$, yielding $10\,101$ discrete settings~\cite{DBLP:conf/ijcci/ThomaserVBK23}. We construct the surrogate model $\fhats$ for the source functions to validate our proposed transfer learning approach on this dataset by utilizing the entire $\fs$ dataset. The complete dataset is also employed as the test set to compute the SMAPE for different GPR models on the target function $\ft$. The transfer dataset $\mathcal{T}$ is randomly sampled from the target function, with a maximum of up to $50$ points. A GPR model is also trained from scratch using only the same transfer dataset without incorporating any prior knowledge. We also sweep $\mathcal{T}$ from 5 to $50$ to study sample‐size effects, and benchmark against an affine‐only transfer baseline~\cite{PanVerLopBac2024transfer}.



\paragraph{Performance measure}
Model accuracy is evaluated via the symmetric mean absolute percentage error (SMAPE)~\cite{Flores1986pragmatic}. 
To ensure robustness and reliability, the transfer learning process is conducted ten times per BBOB function, using a randomly generated beta CDF and an affine transformation in each iteration. Likewise, the real-world application in the automotive industry is repeated ten times to account for variability.

\paragraph{Implementation details}
\Reviewer{The optimisation of rotation matrices, warp parameters, and translations is likely highly non-convex. While the authors use gradient-based or CMA-ES approaches, there is little discussion of local minima risk or whether multiple restarts and regularisations might be necessary. How difficult is the optimisation in practice? A multi-restart approach (as often used when training GP hyperparameters) might help.}
The BBOB functions are accessed via the IOHexperimenter framework~\cite{IOHexperimenter2024}.
We implement a GPR model with a Gaussian kernel with the \texttt{GPy} package.\footnote{\url{https://gpy.readthedocs.io/en/deploy/}} To mitigate the skewness of the function values, a $log$-transformation is applied before training 
the surrogate model for both synthetic and real-world problems.
We use an exponentially decaying learning rate scheduler for the mini-batch gradient descent to minimize the loss function. In addition, a multi-restart strategy is employed to avoid getting stuck in local minima during the optimization.
Hyperparameter tuning for the transfer learning procedure included adjusting the learning rate ($[10^{-3}, 1]$), batch size ([$0.1|\mathcal{T}|, 0.2|\mathcal{T}|$]), number of epochs ($[60, 100]$), and the decay rate of the exponential scheduler ($[5\times 10^{-3}, 0.3]$). These hyperparameters are independently fine-tuned for each BBOB function using the \texttt{SMAC3} library~\cite{LinEggFeu2022smac3}. The experimental setup details, implementations, and a copy of the supplementary materials are available via Zenodo~\cite{Anonymous-supp}.


\section{Experimental Results} \label{sec:results}
\begin{table*}[!htbp]
\centering
\fontsize{9}{10}\selectfont
\setlength{\tabcolsep}{3pt}
\begin{tabular}{c|ccccccc}
    \toprule 
    2D & Original GPR & Train from scratch & Transferred & Train from scratch & Transferred & Train from scratch & Transferred \\
    \cmidrule(lr){3-4} \cmidrule(lr){5-6} \cmidrule(lr){7-8}
    & & \multicolumn{2}{c}{20 samples} &\multicolumn{2}{c}{40 samples} & \multicolumn{2}{c}{80 samples} \\
    \midrule
    F1 & 0.3701 $\pm$ 0.0370 & 0.1080 $\pm$ 0.0467 & \underline{\textbf{0.0592 $\pm$ 0.0397}} & 0.0580 $\pm$ 0.0212 & \underline{0.0687 $\pm$ 0.0240} & 0.0390 $\pm$ 0.0159 & \underline{0.0609 $\pm$ 0.0429} \\
    F2 & 0.1365 $\pm$ 0.0195 & 0.0613 $\pm$ 0.0219 & \underline{\textbf{0.0318 $\pm$ 0.0183}} & 0.0561 $\pm$ 0.0190 & \underline{\textbf{0.0296 $\pm$ 0.0178}} & 0.0442 $\pm$ 0.0156 & \underline{0.0302 $\pm$ 0.0182} \\
    F3 & 0.2686 $\pm$ 0.0504 & 0.0754 $\pm$ 0.0308 & \underline{0.0564 $\pm$ 0.0240} & 0.0584 $\pm$ 0.0202 & \underline{0.0512 $\pm$ 0.0158} & 0.0345 $\pm$ 0.0106 & \underline{0.0488 $\pm$ 0.0195} \\
    F4 & 0.2676 $\pm$ 0.0561 & 0.0675 $\pm$ 0.0255 & \underline{0.0545 $\pm$ 0.0116} & 0.0477 $\pm$ 0.0132 & \underline{0.0602 $\pm$ 0.0289} & \textbf{0.0312 $\pm$ 0.0087} & \underline{0.0533 $\pm$ 0.0217} \\
    F5 & 0.2653 $\pm$ 0.0709 & 0.0419 $\pm$ 0.0188 & \underline{0.0472 $\pm$ 0.0283} & \textbf{0.0244 $\pm$ 0.0127} & \underline{0.0461 $\pm$ 0.0277} & \textbf{0.0153 $\pm$ 0.0122} & \underline{0.0421 $\pm$ 0.0268} \\
    F6 & 0.3651 $\pm$ 0.1498 & 0.0865 $\pm$ 0.1002 & \underline{0.0718 $\pm$ 0.0625} & 0.0751 $\pm$ 0.0784 & \underline{0.0628 $\pm$ 0.0483} & 0.0557 $\pm$ 0.0473 & \underline{0.0626 $\pm$ 0.0500} \\
    F7 & 0.4115 $\pm$ 0.0824 & 0.1615 $\pm$ 0.0626 & \underline{\textbf{0.0777 $\pm$ 0.0357}} & 0.1151 $\pm$ 0.0406 & \underline{\textbf{0.0749 $\pm$ 0.0379}} & 0.0851 $\pm$ 0.0249 & \underline{0.0755 $\pm$ 0.0327} \\
    F8 & 0.3414 $\pm$ 0.0552 & 0.1378 $\pm$ 0.0551 & \underline{0.1134 $\pm$ 0.0544} & 0.1015 $\pm$ 0.0573 & \underline{0.1065 $\pm$ 0.0442} & 0.0709 $\pm$ 0.0409 & \underline{0.0949 $\pm$ 0.0446} \\
    F9 & 0.3564 $\pm$ 0.0595 & 0.1276 $\pm$ 0.0716 & \underline{0.0827 $\pm$ 0.0443} & 0.1102 $\pm$ 0.0706 & \underline{0.0869 $\pm$ 0.0436} & 0.0714 $\pm$ 0.0363 & \underline{0.0804 $\pm$ 0.0347} \\
    F10 & 0.1477 $\pm$ 0.0315 & 0.0677 $\pm$ 0.0261 & \underline{0.0433 $\pm$ 0.0228} & 0.0586 $\pm$ 0.0177 & \underline{0.0447 $\pm$ 0.0226} & 0.0493 $\pm$ 0.0136 & \underline{0.0405 $\pm$ 0.0227} \\
    F11 & 0.1913 $\pm$ 0.0793 & 0.0697 $\pm$ 0.0305 & \underline{0.0513 $\pm$ 0.0265} & 0.0564 $\pm$ 0.0235 & \underline{0.0468 $\pm$ 0.0281} & 0.0469 $\pm$ 0.0138 & \underline{0.0458 $\pm$ 0.0268} \\
    F12 & 0.3242 $\pm$ 0.0535 & 0.0527 $\pm$ 0.0239 & \underline{0.0512 $\pm$ 0.0232} & 0.0418 $\pm$ 0.0156 & \underline{0.0481 $\pm$ 0.0258} & 0.0318 $\pm$ 0.0151 & \underline{0.0486 $\pm$ 0.0252} \\
    F13 & 0.1815 $\pm$ 0.0210 & 0.0601 $\pm$ 0.0198 & \underline{\textbf{0.0373 $\pm$ 0.0151}} & 0.0492 $\pm$ 0.0105 & \underline{0.0398 $\pm$ 0.0223} & 0.0400 $\pm$ 0.0080 & \underline{0.0367 $\pm$ 0.0189} \\
    F14 & 0.5601 $\pm$ 0.0838 & 0.1666 $\pm$ 0.0710 & \underline{0.1117 $\pm$ 0.0678} & 0.1112 $\pm$ 0.0363 & \underline{0.1254 $\pm$ 0.0609} & \textbf{0.0554 $\pm$ 0.0194} & \underline{0.1312 $\pm$ 0.0672} \\
    F15 & 0.3755 $\pm$ 0.1195 & 0.0833 $\pm$ 0.0351 & \underline{0.0627 $\pm$ 0.0330} & 0.0536 $\pm$ 0.0171 & \underline{0.0558 $\pm$ 0.0228} & \textbf{0.0362 $\pm$ 0.0159} & \underline{0.0663 $\pm$ 0.0329} \\
    F16 & 0.2039 $\pm$ 0.0268 & 0.1433 $\pm$ 0.0353 & \underline{0.1348 $\pm$ 0.0304} & 0.1334 $\pm$ 0.0270 & \underline{0.1317 $\pm$ 0.0227} & 0.1246 $\pm$ 0.0256 & \underline{0.1212 $\pm$ 0.0252} \\
    F17 & 0.5333 $\pm$ 0.1150 & 0.1368 $\pm$ 0.0389 & \underline{0.1236 $\pm$ 0.0340} & 0.1215 $\pm$ 0.0363 & \underline{0.1161 $\pm$ 0.0291} & 0.1076 $\pm$ 0.0300 & \underline{0.1161 $\pm$ 0.0294} \\
    F18 & 0.4151 $\pm$ 0.0884 & 0.1047 $\pm$ 0.0372 & \underline{0.0944 $\pm$ 0.0263} & 0.0926 $\pm$ 0.0304 & \underline{0.0936 $\pm$ 0.0277} & 0.0848 $\pm$ 0.0272 & \underline{0.0848 $\pm$ 0.0229} \\
    F19 & 0.4308 $\pm$ 0.1200 & 0.1499 $\pm$ 0.0447 & \underline{0.1407 $\pm$ 0.0486} & 0.1339 $\pm$ 0.0386 & \underline{0.1312 $\pm$ 0.0442} & 0.1111 $\pm$ 0.0343 & \underline{0.1355 $\pm$ 0.0415} \\
    F20 & 0.4005 $\pm$ 0.0336 & 0.1500 $\pm$ 0.0820 & \underline{\textbf{0.0728 $\pm$ 0.0616}} & 0.1097 $\pm$ 0.0588 & \underline{0.0708 $\pm$ 0.0556} & 0.0788 $\pm$ 0.0380 & \underline{0.0691 $\pm$ 0.0505} \\
    F21 & 0.3857 $\pm$ 0.0605 & 0.1972 $\pm$ 0.0439 & \underline{0.1782 $\pm$ 0.0491} & 0.1807 $\pm$ 0.0411 & \underline{0.1728 $\pm$ 0.0401} & 0.1693 $\pm$ 0.0383 & \underline{0.1825 $\pm$ 0.0428} \\
    F22 & 0.2972 $\pm$ 0.0390 & 0.1577 $\pm$ 0.0370 & \underline{0.1445 $\pm$ 0.0405} & 0.1463 $\pm$ 0.0376 & \underline{0.1382 $\pm$ 0.0378} & 0.1312 $\pm$ 0.0294 & \underline{0.1395 $\pm$ 0.0358} \\
    F23 & 0.1543 $\pm$ 0.0122 & 0.1575 $\pm$ 0.0185 & 0.1538 $\pm$ 0.0124 & 0.1556 $\pm$ 0.0168 & 0.1539 $\pm$ 0.0127 & 0.1541 $\pm$ 0.0149 & 0.1538 $\pm$ 0.0128 \\
    F24 & 0.4134 $\pm$ 0.2179 & 0.1821 $\pm$ 0.0660 & 0.3103 $\pm$ 0.1820 & 0.1503 $\pm$ 0.0434 & 0.3081 $\pm$ 0.1839 & \textbf{0.1214 $\pm$ 0.0365} & 0.3060 $\pm$ 0.1857 \\
    \bottomrule
\end{tabular}%
\caption{On 2D BBOB functions, we compare the SMAPE value (mean $\pm$ standard deviation) of three GPR models: the original model, the transferred model, and the one trained from scratch on the transfer dataset. Three transfer sample sizes are investigated $\abs{\mathcal{T}}\in\{20, 40, 80\}$. The transfer target is created with an exponential-shaped beta CDF. We apply the Kruskal-Wallis test with a significance level of 5\%, followed by Dunn's post-hoc analysis to detect significant winners: the transferred model is underlined if it outperforms the original; and the better of the transferred and one trained from scratch is shown in bold.
}
\label{table:finalGPR_2D}
\end{table*}

\subsection{Transferring GPR on BBOB}
In Fig.~\ref{figure:smape_heatmap_2d}, we compare the transferred GPR to the one trained from scratch on each 2-dimensional BBOB function regarding the average SMAPE difference. This analysis investigates the effectiveness of transfer learning with different transfer sample sizes and beta CDF parameterizations. The results reveal that, with only 20 transfer samples, the transferred model performs better for most of the function and beta parameterization combinations. However, as the number of transfer samples increases, the performance advantage decreases until it becomes negative for a sample size of 80. Also, the transfer learning method fails to improve on F24, which has a highly rugged landscape. 

\begin{figure*}[!htbp]
  \centering
  \includegraphics[width=\textwidth]{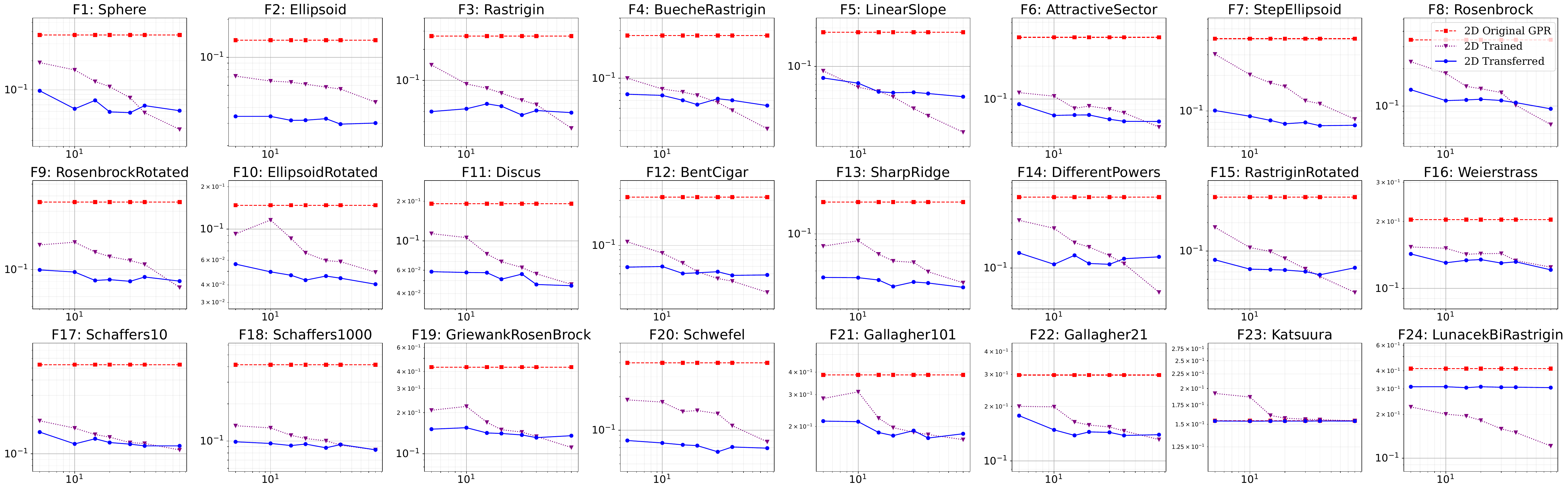}
  \caption{The SMAPE values ($y$-axis) for the original GPR, transferred GPR, and GPR trained solely on the transfer dataset are plotted against the transfer dataset sizes ($x$-axis: 5, 10, 15, 20, 30, 40, 80) for 2D BBOB functions. The analysis combines a beta CDF warping function (approximating an exponential transformation) with an affine transformation.}
  \Description{The SMAPE values (displayed on the $y$-axis) are visualized for three different models: the original GPR, the transferred GPR, and the one trained solely on the transfer dataset.}
  \label{figure:GPRSMAPEplot_exponential_dim2}
\end{figure*}

\begin{figure*}[!ht]
  \centering
  \includegraphics[width=\textwidth,trim=0 32mm 0 0,clip]{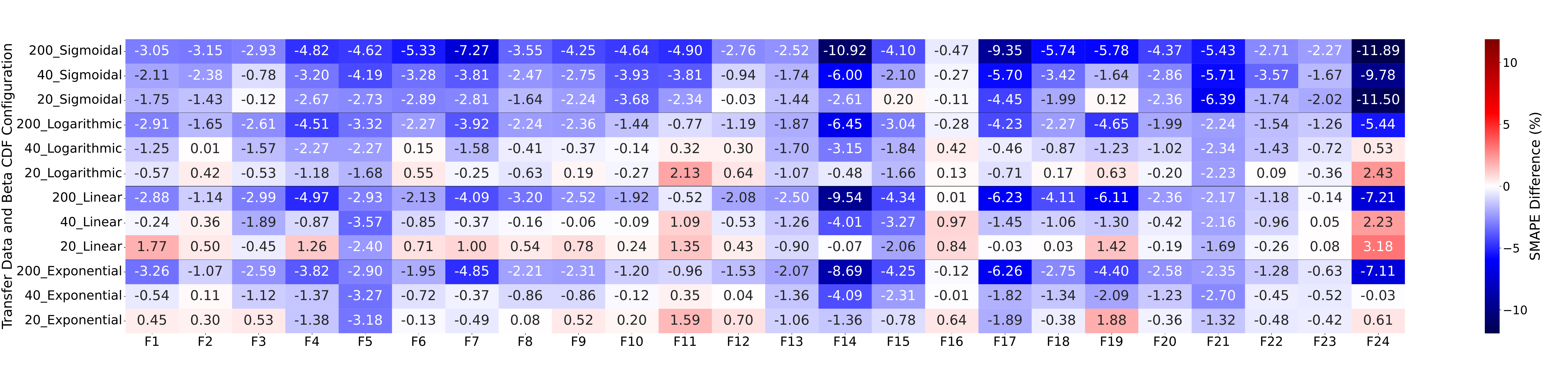}
  \caption{
  On 5D BBOB functions, we compare the transferred GPR models with those trained from scratch on the transfer dataset. Each cell displays the percentage difference in average SMAPE (\%) for a combination of BBOB functions, sample size, and beta CDF shape. Positive values (shown in red) indicate superior performance of the transferred model and vice versa.}
  \Description{Comparison of Gaussian process regression models obtained through transfer learning with those trained from scratch on the transfer dataset, evaluated on 5-dimensional BBOB functions.}
  \label{figure:smape_heatmap_5d}
\end{figure*}

\begin{figure*}[!ht]
  \centering
  \includegraphics[width=\textwidth,trim=0 32mm 0 0,clip]{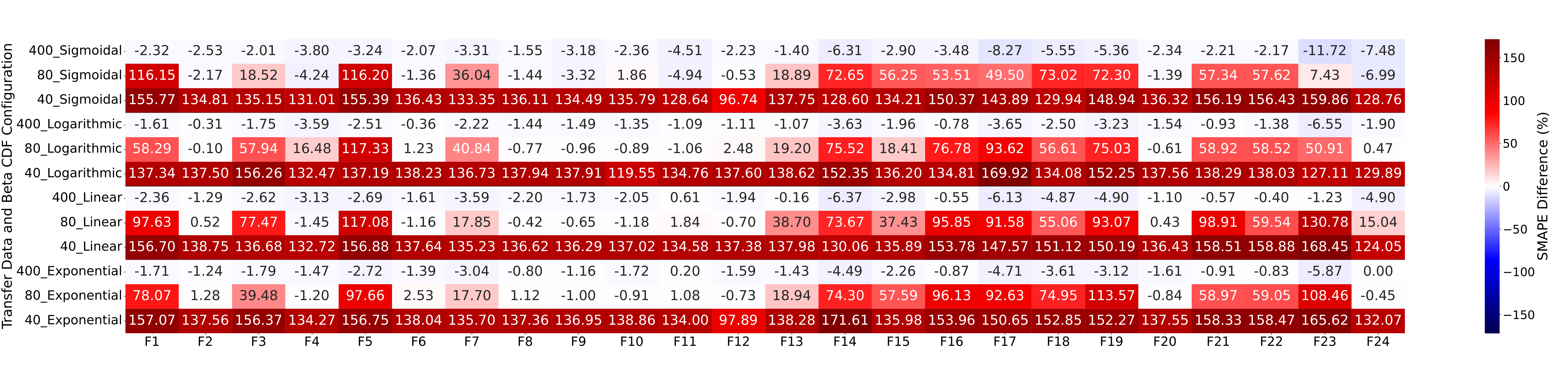}
  \caption{
  On 10D BBOB functions, we compare the transferred GPR models with those trained from scratch on the transfer dataset. Each cell displays the percentage difference in average SMAPE (\%) for a combination of BBOB functions, sample size, and beta CDF shape. Positive values (shown in red) indicate superior performance of the transferred model and vice versa.}
  \Description{Comparison of Gaussian process regression models obtained through transfer learning with those trained from scratch on the transfer dataset, evaluated on 10-dimensional BBOB functions.}
  \label{figure:smape_heatmap_10d}
\end{figure*}

Additional analyses in Fig.~\ref{figure:smape_heatmap_5d} and Fig.~\ref{figure:smape_heatmap_10d} (refer to the supplementary material~\cite{Anonymous-supp}) extend these observations to 5D and 10D BBOB functions.
In the 5D case, models trained from scratch demonstrate a more noticeable advantage compared to the 2D scenario, surpassing transferred models on certain BBOB functions with as few as 20 samples. This could be because the 5D case involves more complex relations, which the transferred models find harder to capture than the simpler or more closely related 2D case.
Interestingly, in the 10D experiments, models trained from scratch exhibit significant performance drops with only 40 samples. A similar issue is observed for some BBOB functions even with 80 samples, as such sample sizes are insufficient to train accurate GPR models~\cite{Bull11, van2008rates}.
These results align with prior studies~\cite{PanVerLopBac2024transfer}, which emphasize the value of transfer learning in data-scarce settings but highlight its diminishing returns as data availability grows.

Next, we show the detailed performance values of the 2D scenario in Table~\ref{table:finalGPR_2D}. 
Results for 5D and 10D cases are in the supplementary material~\cite{Anonymous-supp}. To assess the impact of transfer sample size, we compare models using the Kruskal-Wallis test and Dunn's post-hoc analysis (5\% significance). Significant results are highlighted: the transferred model is underlined when it outperforms the original, and the better model between transferred and trained-from-scratch is highlighted in boldface when statistically significant.

For 2D functions, the transferred GPR generally outperforms the model trained from scratch and the original GPR when only 20 transfer samples are available, with F5 and F24 being notable exceptions. However, as the transfer sample size increases to 80, most functions perform better with the model trained from scratch than the transferred model. Despite this, the transferred model consistently delivers significant improvements over the original GPR across all transfer sample sizes, except for F23 and F24. 
Fig.~\ref{figure:GPRSMAPEplot_exponential_dim2} further explores the relation between model performance and transfer sample sizes. 
The SMAPE trends show that the transferred model consistently outperforms the model trained from scratch up to 40 sample points for most functions, after which the advantage gradually diminishes. This demonstrates the effectiveness of our transfer learning approach in low-data scenarios, as anticipated. Interestingly, for F5, a simple linear function, the model trained from scratch outperforms the transferred model with just 20 samples—likely due to F5's low sample complexity, which allows accurate learning from minimal data.
A detailed analysis reveals that the original GPR model for F23 and F24 suffers from significant underfitting, particularly in the case of F23, which limits the potential benefits of transfer learning. This finding emphasizes that the effectiveness of the transfer learning approach depends on the original GPR model achieving a baseline level of accuracy. Furthermore, the SMAPE trends for F24 highlight specific limitations of the proposed method, indicating that while effective overall, it may struggle with particular functions.

We now analyze how the dimensionality of a function's domain influences the effectiveness of transfer learning. For 5D functions (refer to Fig.~\ref{figure:GPRSMAPEplot_exponential_dim5} and Table~\ref{table:finalGPR_5D} in the supplementary material~\cite{Anonymous-supp}), the transferred model significantly outperforms the original GPR across most BBOB functions, regardless of the transfer sample size. However, exceptions are observed in functions F16, F21, and F23. As the transfer data size increases to 40, a growing number of BBOB functions favor models trained from scratch, reducing the relative benefits of transfer learning compared to the 2D scenario. When the transfer sample size reaches 200, the model trained from scratch significantly outperforms the transferred model on several functions. For instance, on F5, the trained-from-scratch model consistently delivers superior performance, regardless of sample size. On F16 and F23 both methods remain indistinguishable at all sizes, suggesting these landscapes are simply hard for GPR to capture.

\Reviewer{Regarding Table 3 in the supplementary material: While it makes sense to me that 10D instances are hard
to learn with just 40 or 80 samples, I still wonder why the original GPR model performs considerably better.
Can you perhaps discuss the reason for that? Are the transformations
used by BBOB so mild that a model that has the right shape, but no
data from the target instance, is still reasonably accurate?}

For 10D functions (refer to the supplementary material~\cite{Anonymous-supp}), models trained from scratch struggle with poor performance when sample sizes are small. As dimensionality increases, the complexity of the function landscape requires substantially more data for GPR models to achieve accurate approximations, as previously noted. Under such data-scarce conditions, transfer learning provides a clear advantage over training from scratch. Interestingly, a significant phase transition occurs around 80 samples for the scratch-trained model, marked by a sharp drop in its SMAPE value. This significantly narrows the performance gap with the transferred model. At the largest tested sample size (400), the scratch-trained model outperforms the transferred model across most functions. However, for highly multimodal functions like F16 and F23, transfer learning fails to improve the performance of the original GPR model. In contrast, training the GPR model from scratch with 400 samples achieves superior results on these complex functions. This underscores the limitations of transfer learning in handling highly intricate and multimodal landscapes, where large amounts of task-specific data are essential for optimal performance. In addition, it is noticeable that the original GPR model still outperforms the model trained from scratch on just 40 or 80 samples, despite never seeing target-instance data. This suggests that the original model's fit to the 10-dimensional BBOB landscapes, learned from sufficient source data, remains relatively good even without further adaptation.


\subsection{Ablation study of transferring GPR on BBOB}

\begin{figure*}[!htbp]
  \centering
  \includegraphics[width=\textwidth]{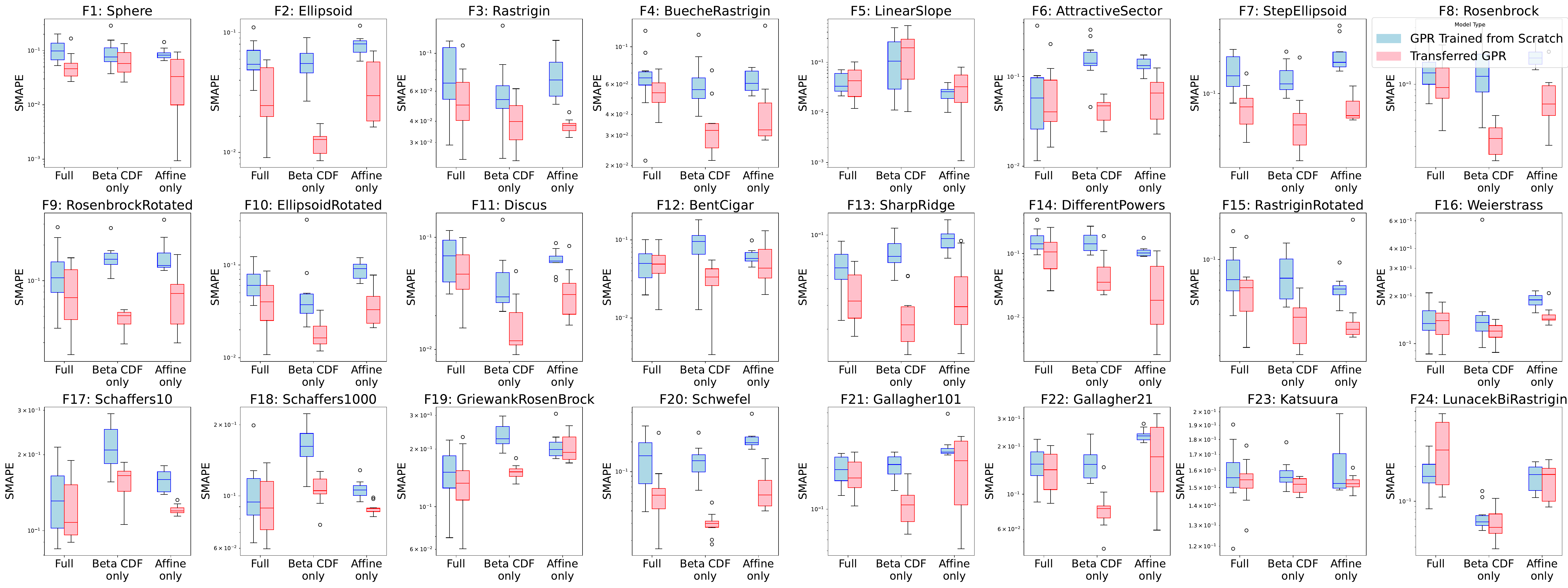}
  \caption{The ablation study focuses on the beta CDF warping function, with rotation and translation disabled, approximating an exponential transformation. We compare our results with reproduced code from~\cite{PanVerLopBac2024transfer} using box plots for 2D BBOB functions with a 20-sample transfer dataset. The plots show SMAPE values ($y$-axis) for the transferred GPR and a model trained solely on the transfer dataset across different transfer learning settings ($x$-axis).}
  \Description{Ablation study focuses exclusively on the beta CDF warping function, with rotation and translation disabled, where the beta CDF approximates an exponential transformation.}
  \label{figure: GPRSMAPEplot_ablation_betacdf_only_exponential_dim2}
\end{figure*}

\begin{figure*}[!htbp]
  \centering
  \includegraphics[width=\textwidth]{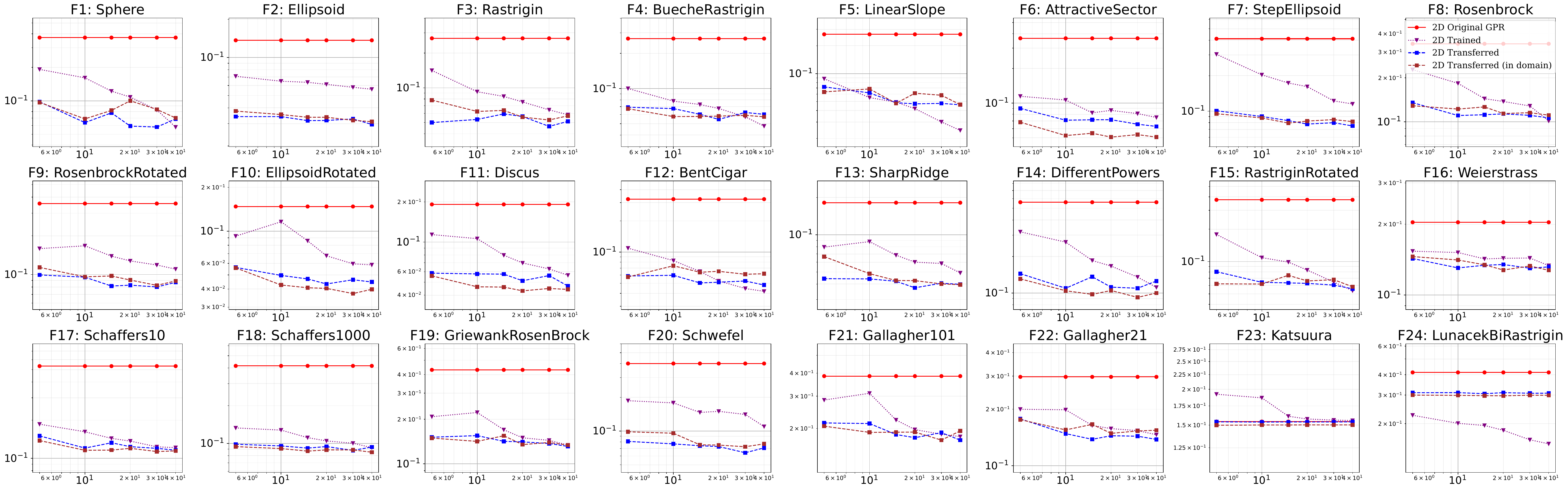}
  \caption{Ablation study presents results for the ``in domain'' scenario, where only transfer data—sampled from the target domain and mapped back into the original domain after transformation—is used for training. SMAPE values ($y$-axis) are shown for the original GPR, transferred GPR, and a model trained solely on the transfer dataset, plotted against transfer dataset sizes ($x$-axis: 5, 10, 15, 20, 30, 40) for 2D BBOB functions. The analysis combines a beta CDF warping function (approximating an exponential transformation) with an affine transformation.}
  \Description{Ablation study. This figure includes results for a scenario where only the transfer data—sampled directly from the target domain and mapped back into the original domain after transformation (referred to as ``in domain'')—is used for training.}
  \label{figure: GPRSMAPEplot_ablation_in_domain_data_dim2}
\end{figure*}

\begin{figure*}[!htbp]
  \centering
  \includegraphics[width=\textwidth]{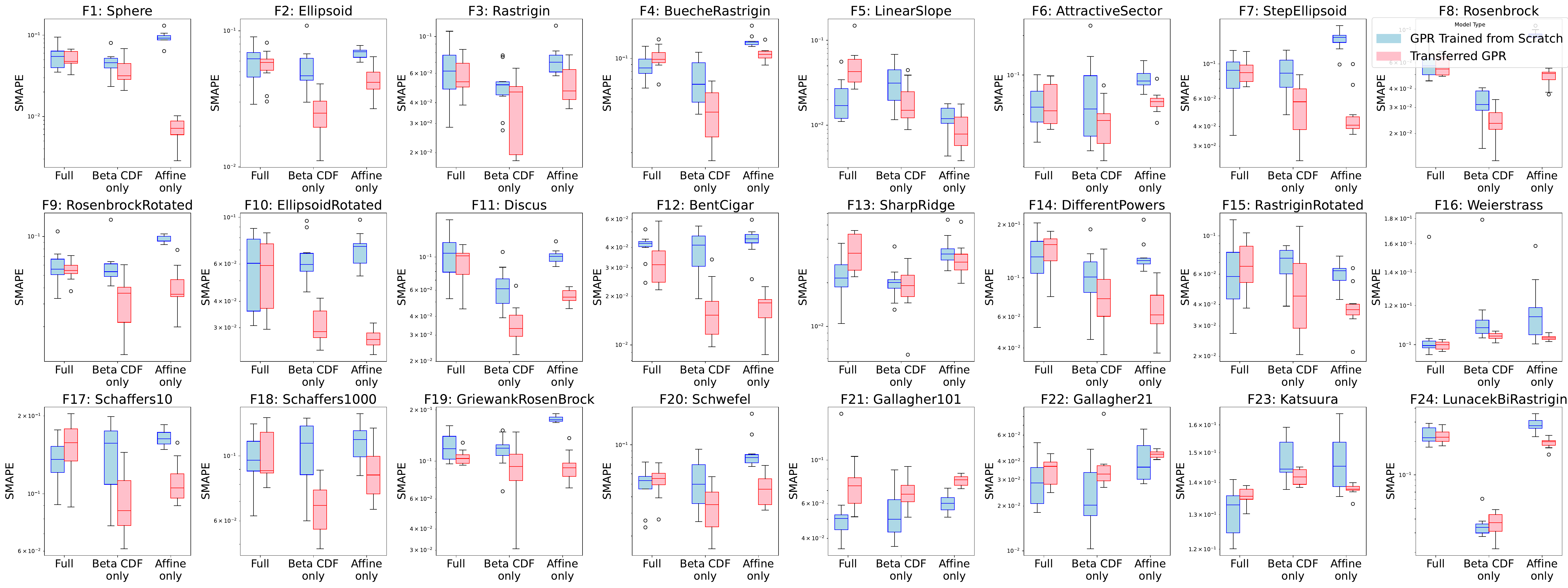}
  \caption{The ablation study focuses on the beta CDF warping function, with rotation and translation disabled, approximating an exponential transformation. We compare our results with reproduced code from~\cite{PanVerLopBac2024transfer} using box plots for 5D BBOB functions with a 20-sample transfer dataset. The plots show SMAPE values ($y$-axis) for the transferred GPR and a model trained solely on the transfer dataset across different transfer learning settings ($x$-axis).}
  \Description{Ablation study focuses exclusively on the beta CDF warping function, with rotation and translation disabled, where the beta CDF approximates an exponential transformation.}
  \label{figure: GPRSMAPEplot_ablation_betacdf_only_exponential_dim5}
\end{figure*}

Fig.~\ref{figure: GPRSMAPEplot_ablation_betacdf_only_exponential_dim2} shows an ablation study on 2D BBOB functions, analyzing the impact of using only the beta CDF warping function, approximating an exponential transformation without rotation or translation. Results are compared to reproduced code from~\cite{PanVerLopBac2024transfer}, which uses only affine transformations. The performance is visualized using box plots of raw SMAPE values obtained from 10 repetitions on 2D BBOB functions, with a transfer dataset of 20 samples. These plots compare the transferred GPR and a model trained solely on the transfer dataset. Since the target functions differ across the three settings, the box plot distributions highlight that, for most BBOB functions, optimizing only the beta CDF parameters in the transferred GPR outperforms jointly optimizing beta CDF and affine parameters and optimizing affine parameters alone.
This highlights that optimizing additional parameters (``Full'') increases the complexity and difficulty of the problem. Moreover, the transferred model consistently outperforms the model trained from scratch when optimizing only the beta CDF parameters across nearly all BBOB functions. 
Fig.~\ref{figure: GPRSMAPEplot_ablation_betacdf_only_exponential_dim5} (refer to the supplementary material~\cite{Anonymous-supp}) presents the results for the 5D case, which exhibits a similar pattern to the observations described above.

The inclusion of rotation and translation in the transformations can introduce boundary effects, where parts of the function landscape initially outside the domain are mapped into it after transformation~\cite{PanVerLopBac2024transfer}. We also examine the "in-domain" setting, where only transfer data—mapped back to the original domain post-transformation—was used for training. This analysis, presented in Fig.~\ref{figure: GPRSMAPEplot_ablation_in_domain_data_dim2}, complements the previously examined scenario of random sampling directly from the target function. This figure demonstrates how model performance varies with different transfer dataset sizes and sampling strategies. For most functions, both ``2D Transferred'' and ``2D Transferred (in the domain)'' exhibit decreasing SMAPE values as the number of samples increases, indicating improved performance with more data. Notably, certain functions, such as F6, F10, F11, and even the more challenging ones like F23 and F24, display significant performance gaps between the two models, emphasizing the benefits of staying within the domain. In contrast, for functions such as F2 and F7, the performance of ``2D Transferred'' and ``2D Transferred (in domain)'' is nearly identical, suggesting that domain restriction has minimal impact in these cases. Although some functions perform better on ``Transferred (in domain)'' than ``Transferred'' under specific transfer dataset sizes, Fig.~\ref{figure: GPRSMAPEplot_ablation_in_domain_data_dim5} (see supplementary material~\cite{Anonymous-supp}) shows that in the 5D case, the difference is minimal, unlike in 2D cases.

\subsection{Transferring GPR on real-world benchmark from automobile industry}

\begin{figure}[t]
  \centering
  \includegraphics[width=0.45\textwidth, trim=0 95 0 90, clip]{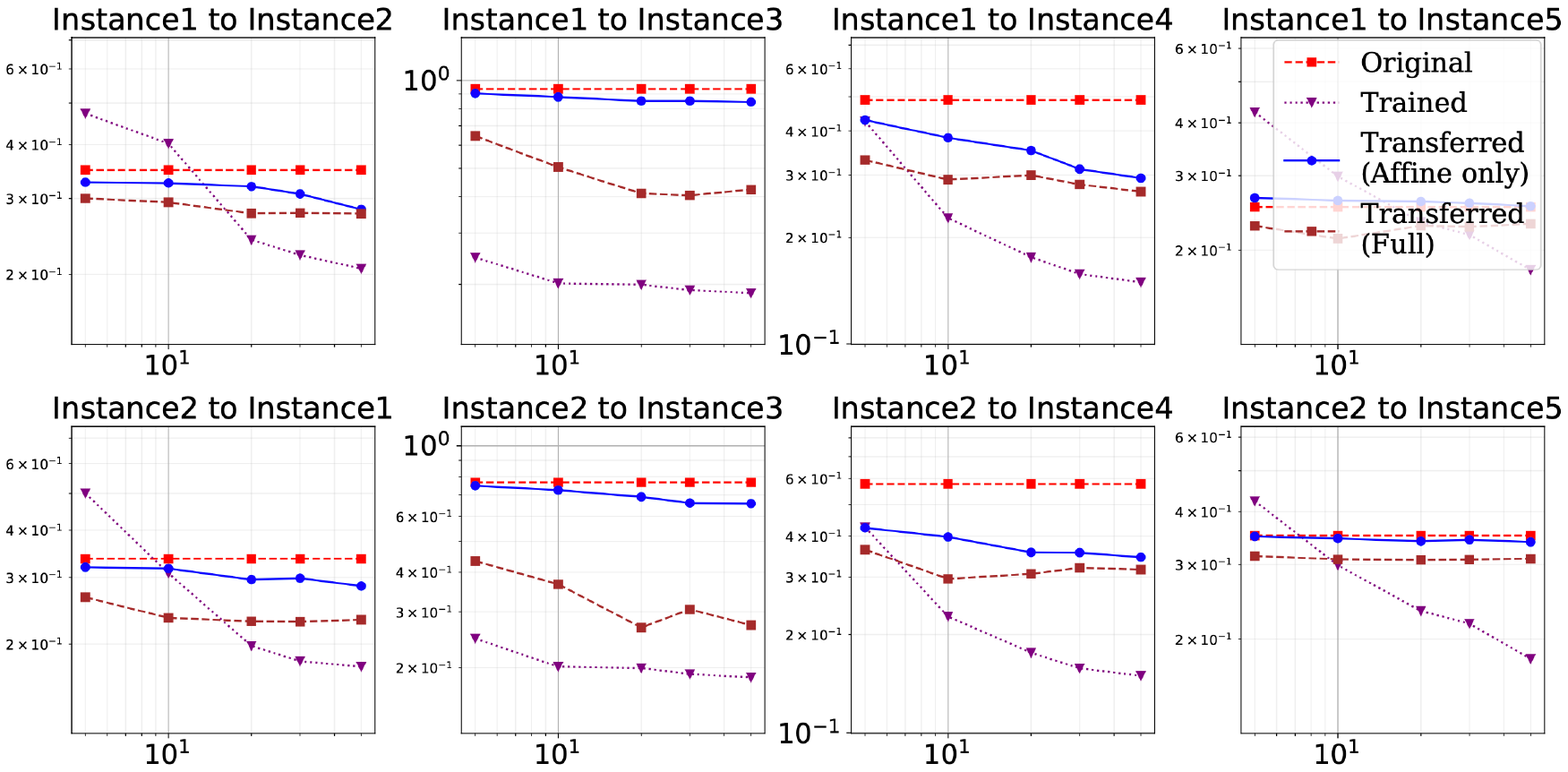}
  \caption{The study evaluates SMAPE values ($y$-axis) for four GPR models on an automotive industry benchmark (Part of): the original GPR model, a transferred GPR with an assumed affine transformation (``Transferred (Affine only)''~\cite{PanVerLopBac2024transfer}), a transferred GPR model using the proposed method (``Transferred (Full)''), and a model trained solely on the transfer dataset. SMAPE values are plotted against transfer dataset sizes ($x$-axis: 5, 10, 20, 30, and 50).}
  \Description{The study evaluates SMAPE values (displayed on the $y$-axis) for four different Gaussian process regression variants on a real-world benchmark within the automotive industry.}
  \label{figure:GPR_SMAPE_plot_real_world_Instance1}
\end{figure}

As highlighted in~\cite{pan2025transferlearningsurrogatemodels}, optimizing with affine transformations has proven effective for many real-world transfer learning applications. However, transferring knowledge between problem instances in the automobile industry presents persistent challenges that require further investigation. We choose this highly challenging benchmark to showcase and evaluate the effectiveness of our transfer learning approach. Fig.~\ref{figure:GPR_SMAPE_plot_real_world_Instance1} (with full results available in Fig.~\ref{figure:GPR_SMAPEplot_real_world_vehicle_2D}, as detailed in the supplementary material~\cite{Anonymous-supp}) illustrates the SMAPE trends for four GPR models across varying transfer dataset sizes, highlighting a subset of the experimental results. These include the original GPR model, a transferred GPR model that assumes an unknown affine transformation between the source and target problem instances (referred to as ``Transferred (Affine only)''~\cite{PanVerLopBac2024transfer}), a transferred GPR model leveraging our proposed method (referred to as ``Transferred (Full)''), and a model trained solely on the transfer dataset. 

Overall, in most cases, the transferred GPR model surpasses the performance of the model trained from scratch, mainly when the transfer dataset is relatively small (fewer than 30 samples). However, as the sample size increases, the performance of the GPR model trained from scratch progressively catches up. Interestingly, there are specific scenarios, such as transferring related to problem instance3, where the transfer learning approach fails. Combined with the instance landscapes discussed in the original study~\cite{DBLP:conf/ijcci/ThomaserVBK23}, these findings suggest that instance3 differs significantly from the other instances, presenting substantial challenges for effective transfer.

The results demonstrate that our proposed transfer learning method consistently outperforms the affine-only transferred approach across various transfer dataset sizes, particularly in scenarios like transferring from instance1 to instance2. Moreover, the proposed method excels in transfers involving instance3, significantly outperforming the affine-only approach, demonstrating its ability to capture more complex relations between source and target functions. However, the scratch-trained model remains the top performer among all GPR variants, indicating that the relations involving instance3 are still too intricate to fully capture.

In the BBOB problem suite, our target functions are explicitly designed so that a perfect transformation exists—meaning that if we replicate the settings used to generate the source problem, the resulting model performs equally well on the target problem as it did on the source (disregarding stochastic variations in sampled evaluation points and the influence of out-of-domain samples). However, real-world scenarios rarely offer such guarantees; we typically lack knowledge of any explicit relation between source and target, let alone whether it aligns with our parameterized transformation space. Consequently, there is a tradeoff between the complexity of the transformation and its optimization feasibility, as partially illustrated by our ablation experiment in Fig.~\ref{figure: GPRSMAPEplot_ablation_betacdf_only_exponential_dim2}.
Additionally, the original GPR model rarely outperforms the transferred or scratch-trained models, especially when the transfer dataset is extremely small (e.g., five samples). However, occasional exceptions do occur. In such cases, the transferred model may overfit the limited data during optimization, resulting in poor generalization to the broader target domain.

\section{Conclusion}\label{sec:conclusion}

We present a transfer learning approach to dealing with a complex, nonlinear covariant shift between the source and target problems. We parameterize the unknown covariant shift as the composition of input warping (implemented with beta CDF) and an affine transformation.
The method leverages a small transfer dataset drawn on the target problem to learn the covariant shift, enabling an effective surrogate model transfer between problems.

Experiments with BBOB functions demonstrate the effectiveness of the proposed method. With 20-sample transfer datasets, the transferred GPR outperforms models trained from scratch, particularly in 10D settings. However, with more samples, especially in 5D, scratch-trained models eventually surpass transferred models. The benefits of transfer learning are limited for highly complex functions like F16 and F21–F24, where the original GPR struggled with accurate approximation. The proposed transfer learning method is also validated on a highly challenging real-world automotive task, demonstrating its effectiveness in low-data scenarios. However, substantial gaps between the source and target domains limit the transfer's effectiveness, even with larger transfer datasets. This highlights the ongoing challenge of adapting the method to different problem instances within the automotive domain and the need for a deeper understanding of their relations. 

Future research includes (1) extending this approach to other regression models, such as random forests; (2) integrating active learning into the transfer learning process to sample data points, which will maximally increase the performance; (3) incremental learning can also be considered that each new sample can be used to fit the previously transferred model; (4) investigating the trade-off between the expressivity of the transformation and the trainability of the transfer learning method; (5) introduce a unified penalty framework that discourages any mapping outside the valid regions of both source and target distributions in high-dimensional or real-world settings; (6) by integrating our transferred surrogate into a standard surrogate‐assisted optimizer, we can analyze whether the transfer learning not only yields a more accurate surrogate but also accelerates the overall optimization process; (7) Kumaraswamy warping~\cite{cowen2022hebo} may offer a faster, more flexible alternative to the beta CDF in practice and is worth investigating.



\bibliographystyle{ACM-Reference-Format}
\bibliography{sample-base,bib/abbrev,bib/journals,bib/authors,bib/articles,bib/biblio,bib/crossref}

\clearpage

\begin{table*}[!ht]
\centering
\fontsize{9}{10}\selectfont
\setlength{\tabcolsep}{3pt}
\begin{tabular}{c|ccccccc}
    \toprule 
    5D & Original GPR & Train from scratch & Transferred & Train from scratch & Transferred & Train from scratch & Transferred \\
    \cmidrule(lr){3-4} \cmidrule(lr){5-6} \cmidrule(lr){7-8}
    & & \multicolumn{2}{c}{20 samples} & \multicolumn{2}{c}{40 samples} & \multicolumn{2}{c}{200 samples} \\
    \midrule
    F1 & 0.1499 $\pm$ 0.0409 & 0.0562 $\pm$ 0.0186 & \underline{0.0517 $\pm$ 0.0112} & 0.0418 $\pm$ 0.0112 & \underline{0.0472 $\pm$ 0.0123} & \textbf{0.0186 $\pm$ 0.0057} & \underline{0.0512 $\pm$ 0.0125} \\
    F2 & 0.1058 $\pm$ 0.0137 & 0.0592 $\pm$ 0.0168 & \underline{0.0562 $\pm$ 0.0148} & 0.0545 $\pm$ 0.0165 & \underline{0.0534 $\pm$ 0.0166} & 0.0340 $\pm$ 0.0081 & \underline{0.0447 $\pm$ 0.0126} \\
    F3 & 0.1165 $\pm$ 0.0337 & 0.0637 $\pm$ 0.0218 & \underline{0.0584 $\pm$ 0.0134} & 0.0416 $\pm$ 0.0154 & \underline{0.0528 $\pm$ 0.0087} & \textbf{0.0218 $\pm$ 0.0068} & \underline{0.0477 $\pm$ 0.0047} \\
    F4 & 0.1924 $\pm$ 0.0537 & 0.0882 $\pm$ 0.0188 & \underline{0.1020 $\pm$ 0.0197} & 0.0784 $\pm$ 0.0180 & \underline{0.0921 $\pm$ 0.0126} & \textbf{0.0434 $\pm$ 0.0075} & \underline{0.0816 $\pm$ 0.0162} \\
    F5 & 0.0865 $\pm$ 0.0284 & \textbf{0.0226 $\pm$ 0.0134} & \underline{0.0544 $\pm$ 0.0340} & \textbf{0.0131 $\pm$ 0.0079} & \underline{0.0458 $\pm$ 0.0139} & \textbf{0.0056 $\pm$ 0.0033} & \underline{0.0346 $\pm$ 0.0072} \\
    F6 & 0.1374 $\pm$ 0.0290 & 0.0611 $\pm$ 0.0226 & \underline{0.0624 $\pm$ 0.0222} & 0.0496 $\pm$ 0.0254 & \underline{0.0568 $\pm$ 0.0241} & \textbf{0.0340 $\pm$ 0.0190} & \underline{0.0535 $\pm$ 0.0192} \\
    F7 & 0.1952 $\pm$ 0.0362 & 0.0860 $\pm$ 0.0262 & \underline{0.0909 $\pm$ 0.0163} & 0.0802 $\pm$ 0.0227 & \underline{0.0839 $\pm$ 0.0204} & \textbf{0.0387 $\pm$ 0.0101} & \underline{0.0872 $\pm$ 0.0138} \\
    F8 & 0.1291 $\pm$ 0.0129 & 0.0583 $\pm$ 0.0108 & \underline{0.0575 $\pm$ 0.0087} & 0.0429 $\pm$ 0.0052 & \underline{0.0515 $\pm$ 0.0059} & \textbf{0.0259 $\pm$ 0.0037} & \underline{0.0480 $\pm$ 0.0087} \\
    F9 & 0.1319 $\pm$ 0.0238 & 0.0608 $\pm$ 0.0195 & \underline{0.0556 $\pm$ 0.0093} & 0.0453 $\pm$ 0.0111 & \underline{0.0539 $\pm$ 0.0093} & \textbf{0.0269 $\pm$ 0.0046} & \underline{0.0500 $\pm$ 0.0065} \\
    F10 & 0.1114 $\pm$ 0.0156 & 0.0585 $\pm$ 0.0216 & \underline{0.0565 $\pm$ 0.0197} & 0.0505 $\pm$ 0.0222 & \underline{0.0517 $\pm$ 0.0163} & 0.0356 $\pm$ 0.0142 & \underline{0.0476 $\pm$ 0.0159} \\
    F11 & 0.1504 $\pm$ 0.0191 & 0.1056 $\pm$ 0.0344 & \underline{0.0897 $\pm$ 0.0255} & 0.0889 $\pm$ 0.0245 & \underline{0.0854 $\pm$ 0.0269} & 0.0682 $\pm$ 0.0196 & \underline{0.0778 $\pm$ 0.0285} \\
    F12 & 0.0906 $\pm$ 0.0426 & 0.0407 $\pm$ 0.0072 & \underline{0.0337 $\pm$ 0.0107} & 0.0312 $\pm$ 0.0104 & \underline{0.0308 $\pm$ 0.0070} & 0.0128 $\pm$ 0.0021 & \underline{0.0281 $\pm$ 0.0080} \\
    F13 & 0.0639 $\pm$ 0.0152 & 0.0228 $\pm$ 0.0078 & \underline{0.0334 $\pm$ 0.0093} & \textbf{0.0181 $\pm$ 0.0085} & \underline{0.0317 $\pm$ 0.0068} & \textbf{0.0088 $\pm$ 0.0037} & \underline{0.0295 $\pm$ 0.0073} \\
    F14 & 0.3513 $\pm$ 0.0437 & 0.1279 $\pm$ 0.0457 & \underline{0.1415 $\pm$ 0.0331} & \textbf{0.0976 $\pm$ 0.0347} & \underline{0.1385 $\pm$ 0.0337} & \textbf{0.0465 $\pm$ 0.0145} & \underline{0.1334 $\pm$ 0.0331} \\
    F15 & 0.1704 $\pm$ 0.0337 & 0.0631 $\pm$ 0.0286 & \underline{0.0709 $\pm$ 0.0218} & \textbf{0.0449 $\pm$ 0.0261} & \underline{0.0680 $\pm$ 0.0166} & \textbf{0.0196 $\pm$ 0.0099} & \underline{0.0621 $\pm$ 0.0149} \\
    F16 & 0.0999 $\pm$ 0.0019 & 0.1061 $\pm$ 0.0198 & 0.0997 $\pm$ 0.0020 & 0.0996 $\pm$ 0.0020 & 0.0997 $\pm$ 0.0020 & 0.0985 $\pm$ 0.0021 & 0.0997 $\pm$ 0.0020 \\
    F17 & 0.3530 $\pm$ 0.0759 & 0.1346 $\pm$ 0.0242 & \underline{0.1535 $\pm$ 0.0353} & 0.1189 $\pm$ 0.0219 & \underline{0.1371 $\pm$ 0.0244} & \textbf{0.0724 $\pm$ 0.0093} & \underline{0.1350 $\pm$ 0.0370} \\
    F18 & 0.2331 $\pm$ 0.0617 & 0.0979 $\pm$ 0.0195 & \underline{0.1017 $\pm$ 0.0200} & 0.0802 $\pm$ 0.0142 & \underline{0.0936 $\pm$ 0.0183} & \textbf{0.0585 $\pm$ 0.0077} & \underline{0.0860 $\pm$ 0.0184} \\
    F19 & 0.2491 $\pm$ 0.0486 & 0.1243 $\pm$ 0.0234 & \underline{0.1055 $\pm$ 0.0100} & \textbf{0.0928 $\pm$ 0.0139} & \underline{0.1137 $\pm$ 0.0195} & \textbf{0.0528 $\pm$ 0.0120} & \underline{0.0968 $\pm$ 0.0129} \\
    F20 & 0.1225 $\pm$ 0.0254 & 0.0506 $\pm$ 0.0154 & \underline{0.0542 $\pm$ 0.0130} & 0.0408 $\pm$ 0.0146 & \underline{0.0531 $\pm$ 0.0127} & \textbf{0.0231 $\pm$ 0.0084} & \underline{0.0489 $\pm$ 0.0152} \\
    F21 & 0.0830 $\pm$ 0.0052 & \textbf{0.0598 $\pm$ 0.0380} & 0.0730 $\pm$ 0.0154 & \textbf{0.0412 $\pm$ 0.0130} & \underline{0.0682 $\pm$ 0.0134} & \textbf{0.0393 $\pm$ 0.0124} & \underline{0.0628 $\pm$ 0.0097} \\
    F22 & 0.0430 $\pm$ 0.0028 & 0.0300 $\pm$ 0.0105 & \underline{0.0348 $\pm$ 0.0068} & 0.0299 $\pm$ 0.0154 & \underline{0.0344 $\pm$ 0.0071} & 0.0204 $\pm$ 0.0047 & \underline{0.0332 $\pm$ 0.0064} \\
    F23 & 0.1361 $\pm$ 0.0026 & 0.1314 $\pm$ 0.0070 & 0.1356 $\pm$ 0.0028 & 0.1303 $\pm$ 0.0061 & 0.1355 $\pm$ 0.0028 & 0.1292 $\pm$ 0.0061 & 0.1355 $\pm$ 0.0028 \\
    F24 & 0.4548 $\pm$ 0.1174 & 0.2294 $\pm$ 0.0394 & \underline{0.2233 $\pm$ 0.0292} & 0.2171 $\pm$ 0.0418 & \underline{0.2174 $\pm$ 0.0281} & \textbf{0.1456 $\pm$ 0.0148} & \underline{0.2167 $\pm$ 0.0265} \\
    \bottomrule
\end{tabular}%
\caption{On 5D BBOB functions, we compare the SMAPE value (mean $\pm$ standard deviation) of three GPR models: the original model, the transferred model, and the one trained from scratch on the transfer dataset. Three transfer sample sizes are investigated $\abs{\mathcal{T}}\in\{20, 40, 200\}$. The transfer target is created with an exponential-shaped beta CDF. We apply the Kruskal-Wallis test with a significance level of 5\%, followed by Dunn's posthoc analysis to detect significant winners: the transferred model has underlined if it outperforms the original model; the boldface indicates the better one between the transferred and the model trained from scratch.
}
\label{table:finalGPR_5D}
\end{table*}

\begin{figure*}[!htbp]
  \centering
  \includegraphics[width=\textwidth]{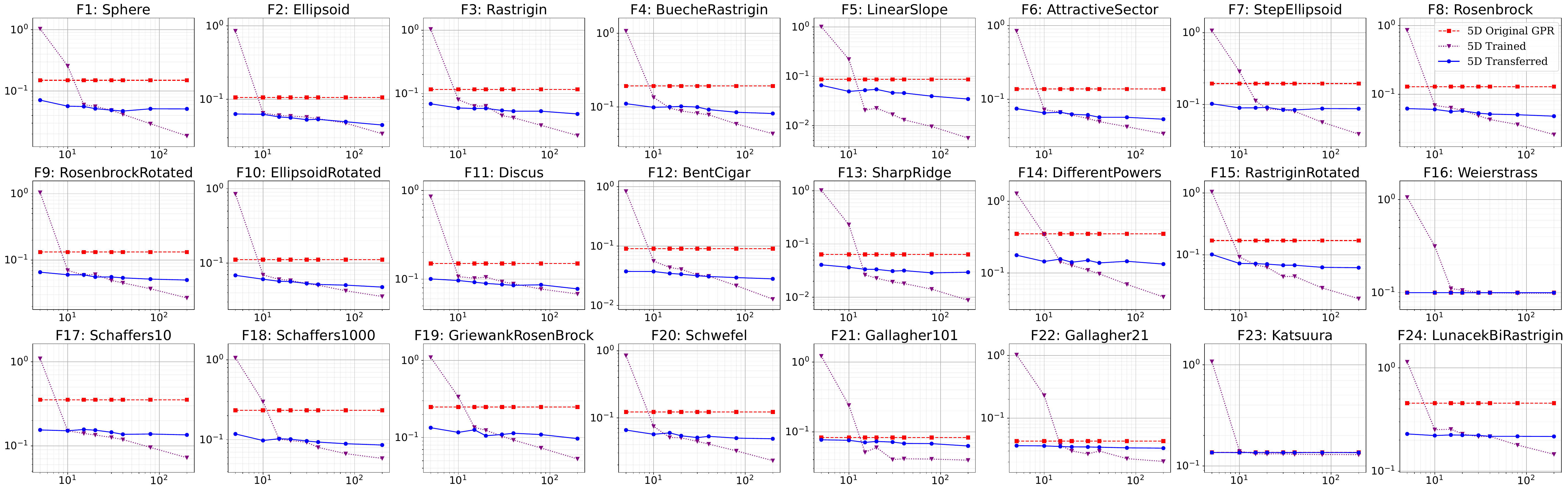}
  \caption{The SMAPE values ($y$-axis) for the original GPR, transferred GPR, and GPR trained solely on the transfer dataset are plotted against the transfer dataset sizes ($x$-axis: 5, 10, 15, 20, 30, 40, 80, 200) for 5D BBOB functions. The analysis combines a beta CDF warping function (approximating an exponential transformation) with an affine transformation.}
  \Description{The SMAPE values (displayed on the $y$-axis) are visualized for three different models: the original GPR, the transferred GPR, and the one trained solely on the transfer dataset.}
  \label{figure:GPRSMAPEplot_exponential_dim5}
\end{figure*}

\begin{table*}[!ht]
\centering
\fontsize{9}{10}\selectfont
\setlength{\tabcolsep}{3pt}
\begin{tabular}{c|ccccccc}
    \toprule 
    10D & Original GPR & Train from scratch & Transferred & Train from scratch & Transferred & Train from scratch & Transferred \\
    \cmidrule(lr){3-4} \cmidrule(lr){5-6} \cmidrule(lr){7-8}
    & & \multicolumn{2}{c}{40 samples} & \multicolumn{2}{c}{80 samples} & \multicolumn{2}{c}{400 samples} \\
    \midrule
    F1 & 0.1116 $\pm$ 0.0286 & 1.6051 $\pm$ 0.7898 & \underline{\textbf{0.0344 $\pm$ 0.0091}} & 0.8132 $\pm$ 0.9690 & \underline{0.0325 $\pm$ 0.0082} & \textbf{0.0139 $\pm$ 0.0033} & \underline{0.0310 $\pm$ 0.0071} \\
    F2 & 0.0812 $\pm$ 0.0139 & 1.4153 $\pm$ 0.8932 & \underline{\textbf{0.0397 $\pm$ 0.0117}} & 0.0494 $\pm$ 0.0326 & \underline{0.0366 $\pm$ 0.0089} & \textbf{0.0226 $\pm$ 0.0080} & \underline{0.0350 $\pm$ 0.0080} \\
    F3 & 0.0887 $\pm$ 0.0219 & 1.6102 $\pm$ 0.7796 & \underline{\textbf{0.0465 $\pm$ 0.0103}} & 0.4379 $\pm$ 0.7816 & \underline{0.0431 $\pm$ 0.0099} & \textbf{0.0234 $\pm$ 0.0053} & \underline{0.0413 $\pm$ 0.0104} \\
    F4 & 0.1661 $\pm$ 0.0531 & 1.4185 $\pm$ 0.8881 & \underline{0.0758 $\pm$ 0.0147} & \textbf{0.0583 $\pm$ 0.0100} & \underline{0.0703 $\pm$ 0.0125} & \textbf{0.0469 $\pm$ 0.0101} & \underline{0.0616 $\pm$ 0.0106} \\
    F5 & 0.0503 $\pm$ 0.0097 & 1.6021 $\pm$ 0.7957 & \underline{\textbf{0.0346 $\pm$ 0.0131}} & 1.0057 $\pm$ 0.9942 & \underline{0.0291 $\pm$ 0.0064} & \textbf{0.0055 $\pm$ 0.0020} & \underline{0.0327 $\pm$ 0.0094} \\
    F6 & 0.0745 $\pm$ 0.0109 & 1.4071 $\pm$ 0.9056 & \underline{\textbf{0.0267 $\pm$ 0.0095}} & 0.0535 $\pm$ 0.0668 & \underline{0.0282 $\pm$ 0.0118} & \textbf{0.0115 $\pm$ 0.0063} & \underline{0.0254 $\pm$ 0.0097} \\
    F7 & 0.1161 $\pm$ 0.0113 & 1.4123 $\pm$ 0.8978 & \underline{\textbf{0.0553 $\pm$ 0.0119}} & 0.2300 $\pm$ 0.5901 & \underline{\textbf{0.0530 $\pm$ 0.0111}} & \textbf{0.0220 $\pm$ 0.0080} & \underline{0.0524 $\pm$ 0.0092} \\
    F8 & 0.0971 $\pm$ 0.0330 & 1.4093 $\pm$ 0.9023 & \underline{\textbf{0.0357 $\pm$ 0.0059}} & 0.0442 $\pm$ 0.0344 & \underline{0.0330 $\pm$ 0.0059} & \textbf{0.0220 $\pm$ 0.0157} & \underline{0.0300 $\pm$ 0.0052} \\
    F9 & 0.0916 $\pm$ 0.0187 & 1.4111 $\pm$ 0.8995 & \underline{\textbf{0.0416 $\pm$ 0.0062}} & \textbf{0.0303 $\pm$ 0.0069} & \underline{0.0403 $\pm$ 0.0048} & \textbf{0.0252 $\pm$ 0.0127} & \underline{0.0368 $\pm$ 0.0030} \\
    F10 & 0.0742 $\pm$ 0.0129 & 1.4260 $\pm$ 0.8782 & \underline{\textbf{0.0374 $\pm$ 0.0098}} & \textbf{0.0275 $\pm$ 0.0106} & \underline{0.0366 $\pm$ 0.0089} & \textbf{0.0176 $\pm$ 0.0065} & \underline{0.0348 $\pm$ 0.0086} \\
    F11 & 0.1458 $\pm$ 0.0088 & 1.4323 $\pm$ 0.8671 & \underline{\textbf{0.0923 $\pm$ 0.0206}} & 0.0946 $\pm$ 0.0177 & \underline{0.0838 $\pm$ 0.0186} & 0.0775 $\pm$ 0.0286 & \underline{0.0755 $\pm$ 0.0175} \\
    F12 & 0.0799 $\pm$ 0.0145 & 1.0147 $\pm$ 0.9853 & \underline{0.0358 $\pm$ 0.0050} & \textbf{0.0279 $\pm$ 0.0029} & \underline{0.0352 $\pm$ 0.0054} & \textbf{0.0163 $\pm$ 0.0026} & \underline{0.0322 $\pm$ 0.0047} \\
    F13 & 0.0359 $\pm$ 0.0055 & 1.4040 $\pm$ 0.9103 & \underline{0.0212 $\pm$ 0.0035} & 0.2104 $\pm$ 0.5965 & \underline{\textbf{0.0210 $\pm$ 0.0042}} & \textbf{0.0064 $\pm$ 0.0019} & \underline{0.0207 $\pm$ 0.0041} \\
    F14 & 0.2082 $\pm$ 0.0191 & 1.8103 $\pm$ 0.5689 & \underline{\textbf{0.0942 $\pm$ 0.0214}} & 0.8378 $\pm$ 0.9489 & \underline{0.0948 $\pm$ 0.0174} & \textbf{0.0392 $\pm$ 0.0127} & \underline{0.0841 $\pm$ 0.0166} \\
    F15 & 0.1094 $\pm$ 0.0404 & 1.4094 $\pm$ 0.9021 & \underline{0.0496 $\pm$ 0.0093} & 0.6220 $\pm$ 0.9021 & \underline{0.0461 $\pm$ 0.0066} & \textbf{0.0210 $\pm$ 0.0042} & \underline{0.0436 $\pm$ 0.0066} \\
    F16 & 0.0733 $\pm$ 0.0132 & 1.6131 $\pm$ 0.7737 & \textbf{0.0735 $\pm$ 0.0142} & 1.0326 $\pm$ 0.9673 & 0.0713 $\pm$ 0.0077 & \textbf{0.0648 $\pm$ 0.0024} & 0.0735 $\pm$ 0.0144 \\
    F17 & 0.2814 $\pm$ 0.0507 & 1.6151 $\pm$ 0.7698 & \underline{\textbf{0.1086 $\pm$ 0.0220}} & 1.0335 $\pm$ 0.9665 & \underline{0.1072 $\pm$ 0.0180} & \textbf{0.0518 $\pm$ 0.0124} & \underline{0.0989 $\pm$ 0.0157} \\
    F18 & 0.2117 $\pm$ 0.0391 & 1.6123 $\pm$ 0.7754 & \underline{\textbf{0.0838 $\pm$ 0.0185}} & 0.8340 $\pm$ 0.9520 & \underline{0.0845 $\pm$ 0.0157} & \textbf{0.0404 $\pm$ 0.0064} & \underline{0.0765 $\pm$ 0.0133} \\
    F19 & 0.2152 $\pm$ 0.0457 & 1.6156 $\pm$ 0.7687 & \underline{\textbf{0.0929 $\pm$ 0.0084}} & 1.2239 $\pm$ 0.9504 & \underline{0.0882 $\pm$ 0.0063} & \textbf{0.0515 $\pm$ 0.0189} & \underline{0.0827 $\pm$ 0.0064} \\
    F20 & 0.0776 $\pm$ 0.0104 & 1.4099 $\pm$ 0.9014 & \underline{\textbf{0.0344 $\pm$ 0.0059}} & \textbf{0.0246 $\pm$ 0.0062} & \underline{0.0330 $\pm$ 0.0058} & \textbf{0.0147 $\pm$ 0.0032} & \underline{0.0308 $\pm$ 0.0057} \\
    F21 & 0.0328 $\pm$ 0.0083 & 1.6041 $\pm$ 0.7917 & \underline{\textbf{0.0208 $\pm$ 0.0045}} & 0.6102 $\pm$ 0.9098 & \underline{0.0205 $\pm$ 0.0046} & \textbf{0.0107 $\pm$ 0.0007} & \underline{0.0198 $\pm$ 0.0043} \\
    F22 & 0.0258 $\pm$ 0.0068 & 1.6023 $\pm$ 0.7953 & \underline{\textbf{0.0176 $\pm$ 0.0046}} & 0.6077 $\pm$ 0.9114 & \underline{0.0172 $\pm$ 0.0044} & \textbf{0.0084 $\pm$ 0.0007} & \underline{0.0167 $\pm$ 0.0042} \\
    F23 & 0.1566 $\pm$ 0.0218 & 1.8111 $\pm$ 0.5667 & \textbf{0.1549 $\pm$ 0.0203} & 1.2393 $\pm$ 0.9316 & 0.1547 $\pm$ 0.0203 & \textbf{0.0961 $\pm$ 0.0115} & 0.1548 $\pm$ 0.0202 \\
    F24 & 0.5525 $\pm$ 0.0623 & 1.4323 $\pm$ 0.8672 & \underline{\textbf{0.1116 $\pm$ 0.0305}} & 0.1069 $\pm$ 0.0357 & \underline{0.1114 $\pm$ 0.0295} & 0.1081 $\pm$ 0.0290 & \underline{0.1081 $\pm$ 0.0290} \\
    \bottomrule
\end{tabular}%
\caption{
On 10D BBOB functions, we compare the SMAPE value (mean $\pm$ standard deviation) of three GPR models: the original model, the transferred model, and the one trained from scratch on the transfer dataset. Three transfer sample sizes are investigated $\abs{\mathcal{T}}\in\{40, 80, 400\}$. The transfer target is created with an exponential-shaped beta CDF. We apply the Kruskal-Wallis test with a significance level of 5\%, followed by Dunn's posthoc analysis to detect significant winners: the transferred model is underlined if it outperforms the original model; the boldface indicates the better one between the transferred and the model trained from scratch.
}
\label{table:finalGPR_10D}
\end{table*}

\begin{figure*}[!htbp]
  \centering
  \includegraphics[width=\textwidth]{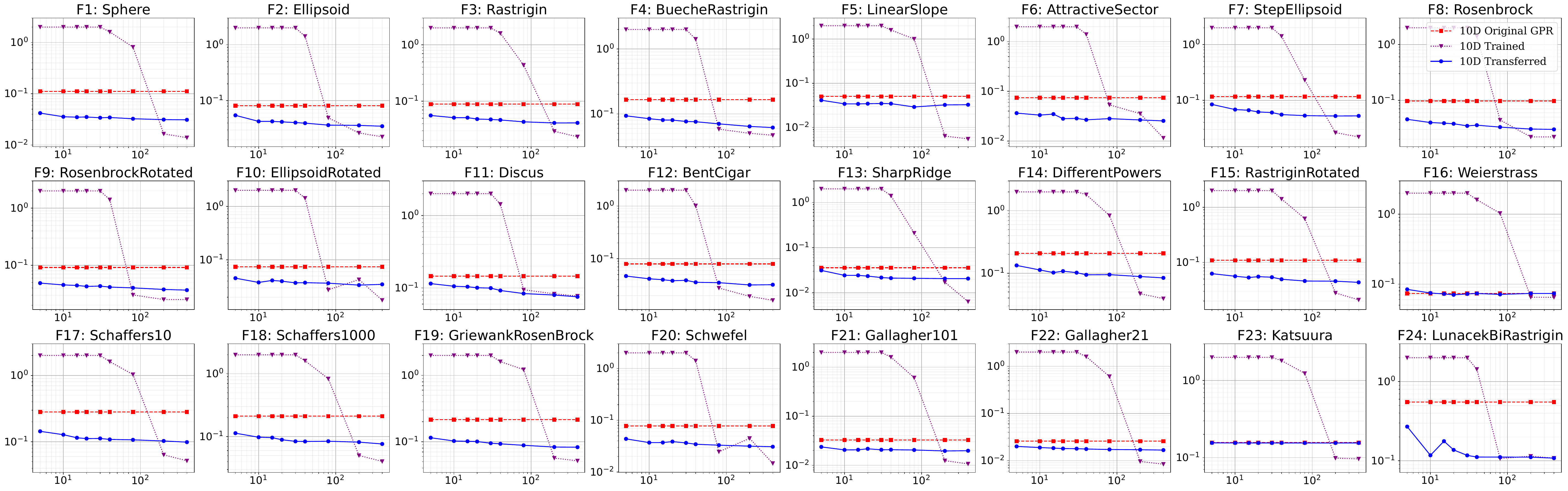}
  \caption{The SMAPE values ($y$-axis) for the original GPR, transferred GPR, and GPR trained solely on the transfer dataset are plotted against the transfer dataset sizes ($x$-axis: 5, 10, 15, 20, 30, 40, 80, 200, 400) for 10D BBOB functions. The analysis combines a beta CDF warping function (approximating an exponential transformation) with an affine transformation.}
  \Description{The SMAPE values (displayed on the $y$-axis) are visualized for three different models: the original GPR, the transferred GPR, and the one trained solely on the transfer dataset.}
  \label{figure:GPRSMAPEplot_exponential_dim10}
\end{figure*}

\begin{figure*}[!htbp]
  \centering
  \includegraphics[width=\textwidth]{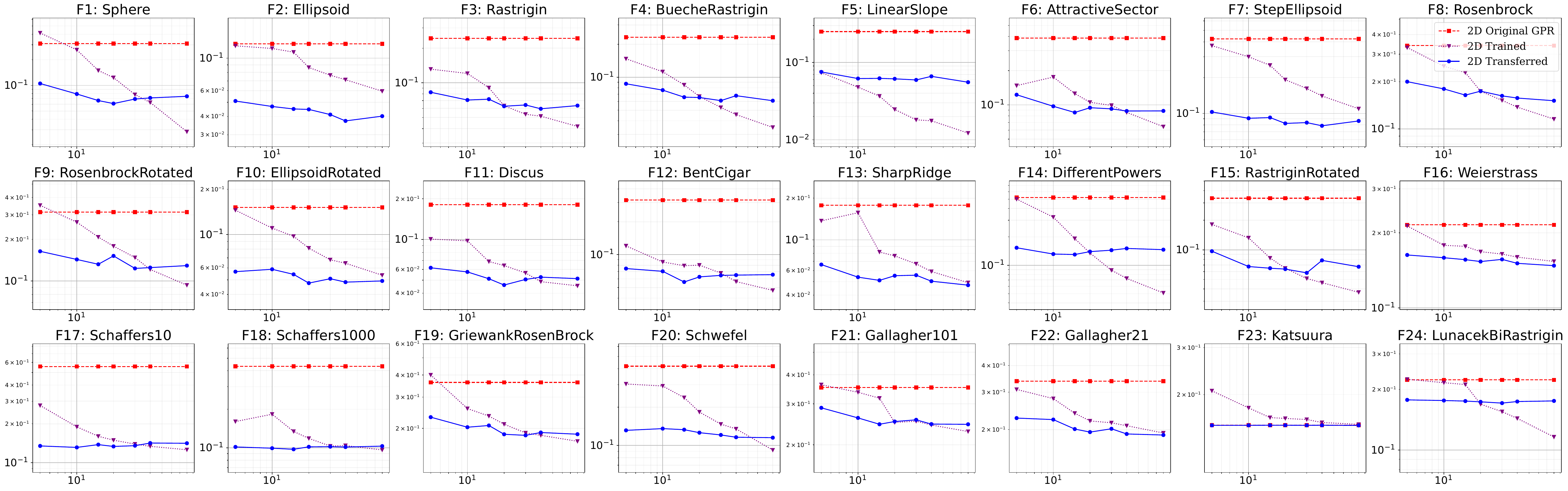}
  \caption{The SMAPE values ($y$-axis) for the original GPR, transferred GPR, and GPR trained solely on the transfer dataset are plotted against the transfer dataset sizes ($x$-axis: 5, 10, 15, 20, 30, 40, 80) for 2D BBOB functions. The analysis combines a beta CDF warping function (approximating a linear transformation) with an affine transformation.}
  \Description{The SMAPE values (displayed on the $y$-axis) are visualized for three different models: the original GPR, the transferred GPR, and the one trained solely on the transfer dataset.}
  \label{figure:GPRSMAPEplot_linear_dim2}
\end{figure*}

\begin{figure*}[!htbp]
  \centering
  \includegraphics[width=\textwidth]{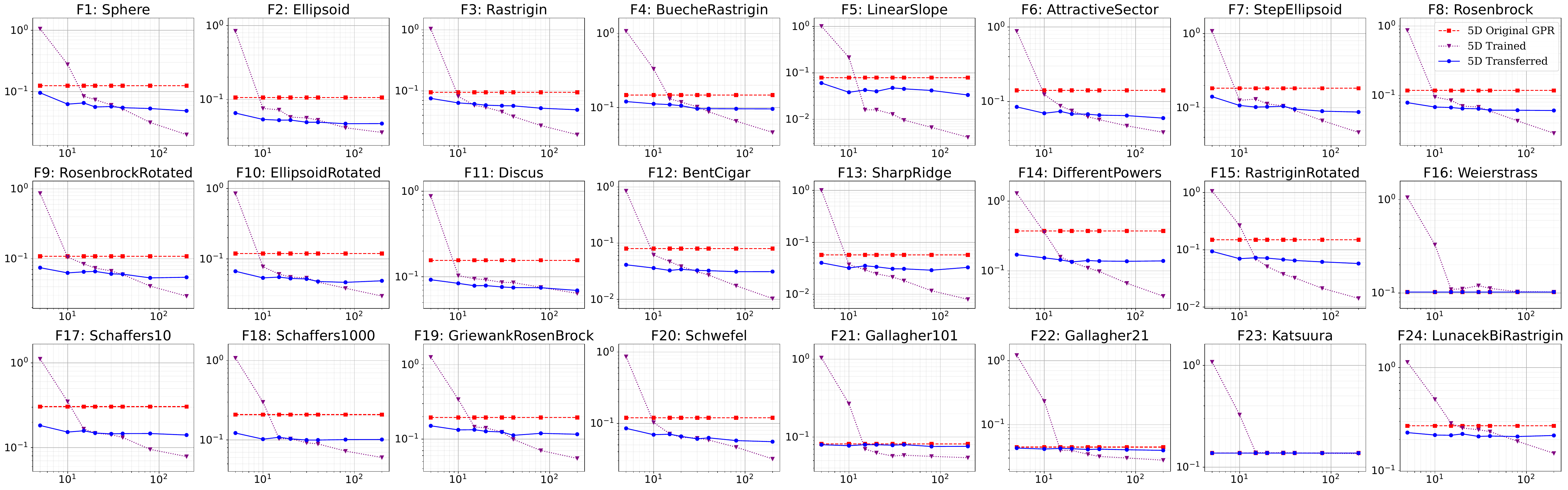}
  \caption{The SMAPE values ($y$-axis) for the original GPR, transferred GPR, and GPR trained solely on the transfer dataset are plotted against the transfer dataset sizes ($x$-axis: 5, 10, 15, 20, 30, 40, 80, 200) for 5D BBOB functions. The analysis combines a beta CDF warping function (approximating a linear transformation) with an affine transformation.}
  \Description{The SMAPE values (displayed on the $y$-axis) are visualized for three different models: the original GPR, the transferred GPR, and the one trained solely on the transfer dataset.}
  \label{figure:GPRSMAPEplot_linear_dim5}
\end{figure*}

\begin{figure*}[!htbp]
  \centering
  \includegraphics[width=\textwidth]{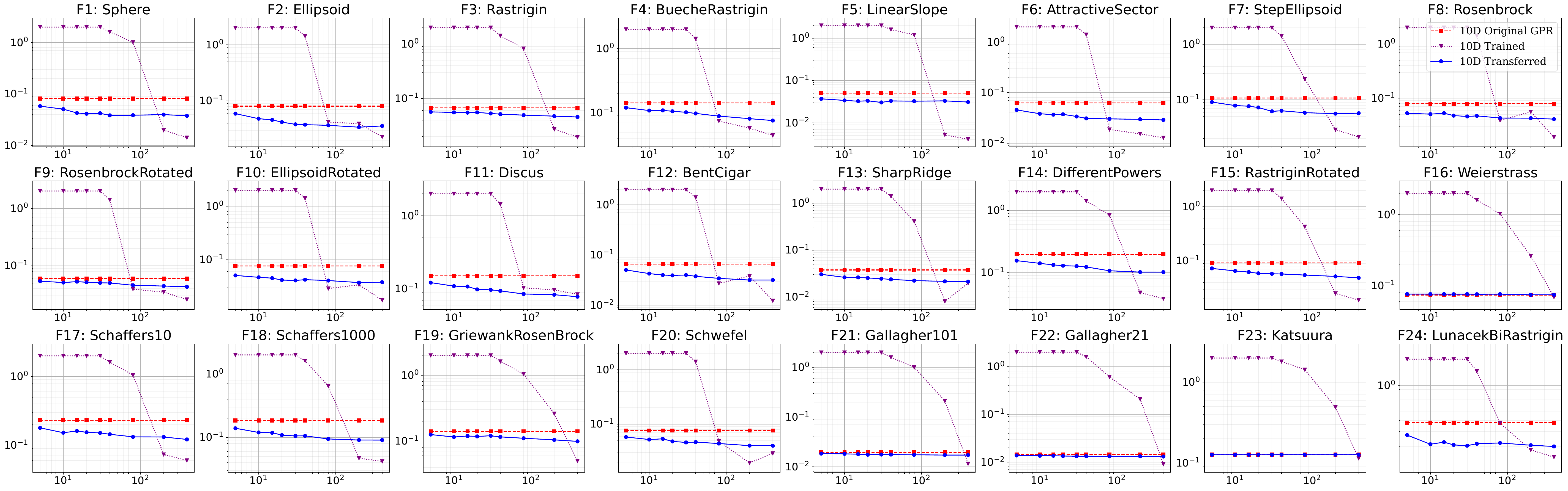}
  \caption{The SMAPE values ($y$-axis) for the original GPR, transferred GPR, and GPR trained solely on the transfer dataset are plotted against the transfer dataset sizes ($x$-axis: 5, 10, 15, 20, 30, 40, 80, 200, 400) for 10D BBOB functions. The analysis combines a beta CDF warping function (approximating a linear transformation) with an affine transformation.}
  \Description{The SMAPE values (displayed on the $y$-axis) are visualized for three different models: the original GPR, the transferred GPR, and the one trained solely on the transfer dataset.}
  \label{figure:GPRSMAPEplot_linear_dim10}
\end{figure*}

\begin{figure*}[!htbp]
  \centering
  \includegraphics[width=\textwidth]{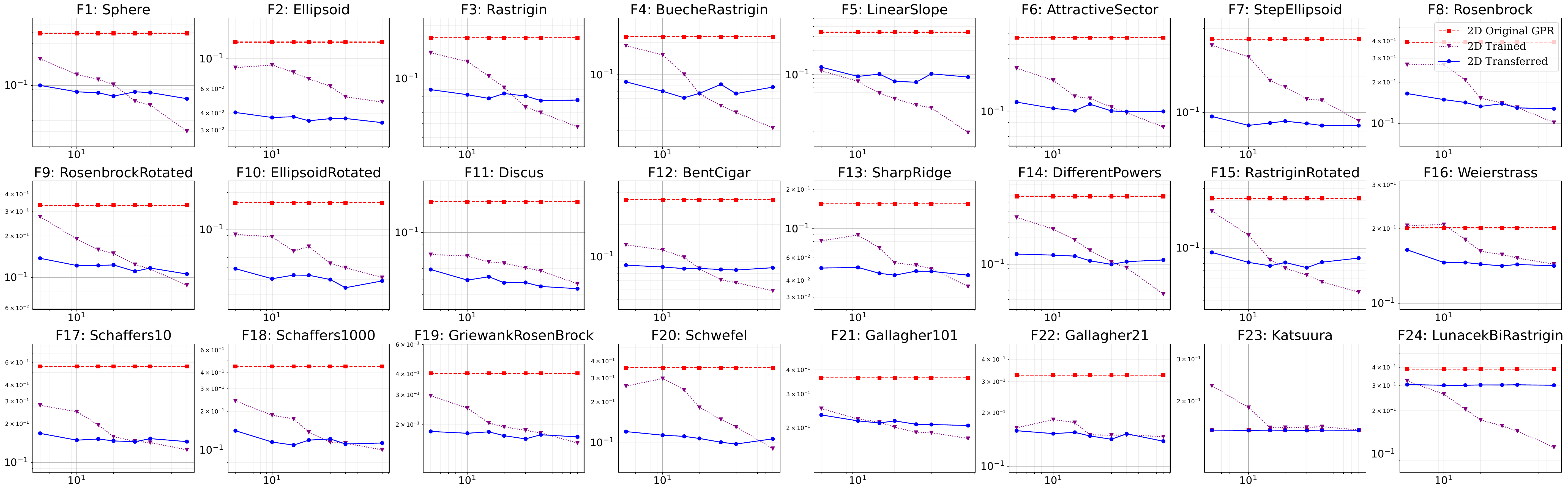}
  \caption{The SMAPE values ($y$-axis) for the original GPR, transferred GPR, and GPR trained solely on the transfer dataset are plotted against the transfer dataset sizes ($x$-axis: 5, 10, 15, 20, 30, 40, 80) for 2D BBOB functions. The analysis combines a beta CDF warping function (approximating a logarithmic transformation) with an affine transformation.}
  \Description{The SMAPE values (displayed on the $y$-axis) are visualized for three different models: the original GPR, the transferred GPR, and the one trained solely on the transfer dataset.}
  \label{figure:GPRSMAPEplot_logarithmic_dim2}
\end{figure*}

\begin{figure*}[!htbp]
  \centering
  \includegraphics[width=\textwidth]{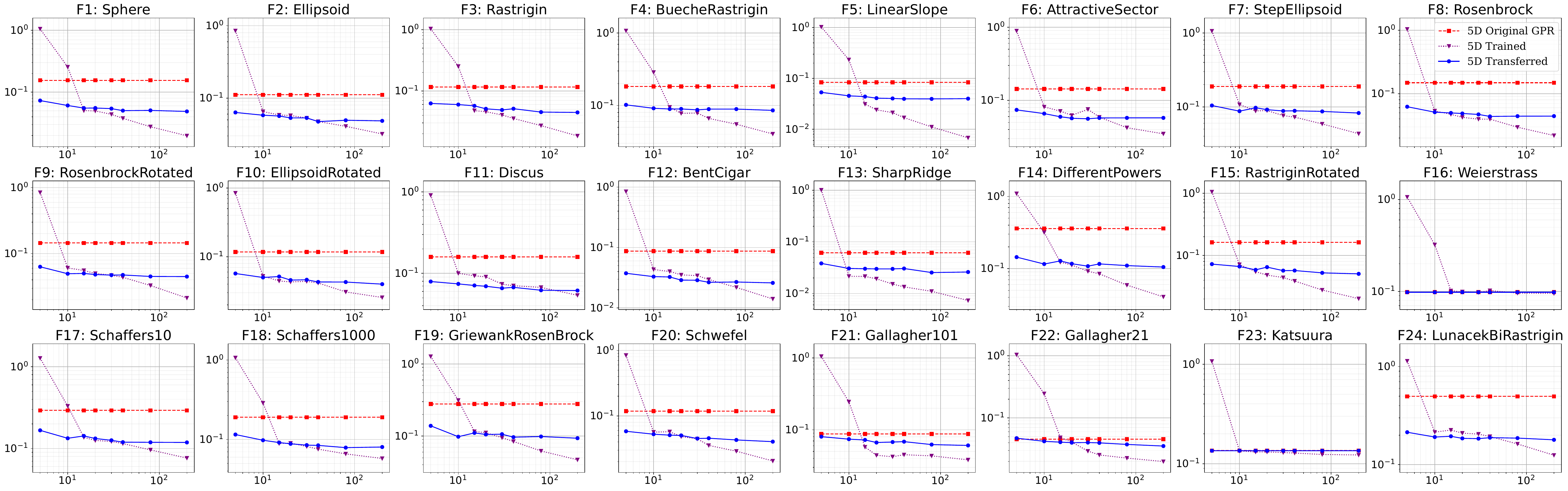}
  \caption{The SMAPE values ($y$-axis) for the original GPR, transferred GPR, and GPR trained solely on the transfer dataset are plotted against the transfer dataset sizes ($x$-axis: 5, 10, 15, 20, 30, 40, 80, 200) for 5D BBOB functions. The analysis combines a beta CDF warping function (approximating a logarithmic transformation) with an affine transformation.}
  \Description{The SMAPE values (displayed on the $y$-axis) are visualized for three different models: the original GPR, the transferred GPR, and the one trained solely on the transfer dataset.}
  \label{figure:GPRSMAPEplot_logarithmic_dim5}
\end{figure*}

\begin{figure*}[!htbp]
  \centering
  \includegraphics[width=\textwidth]{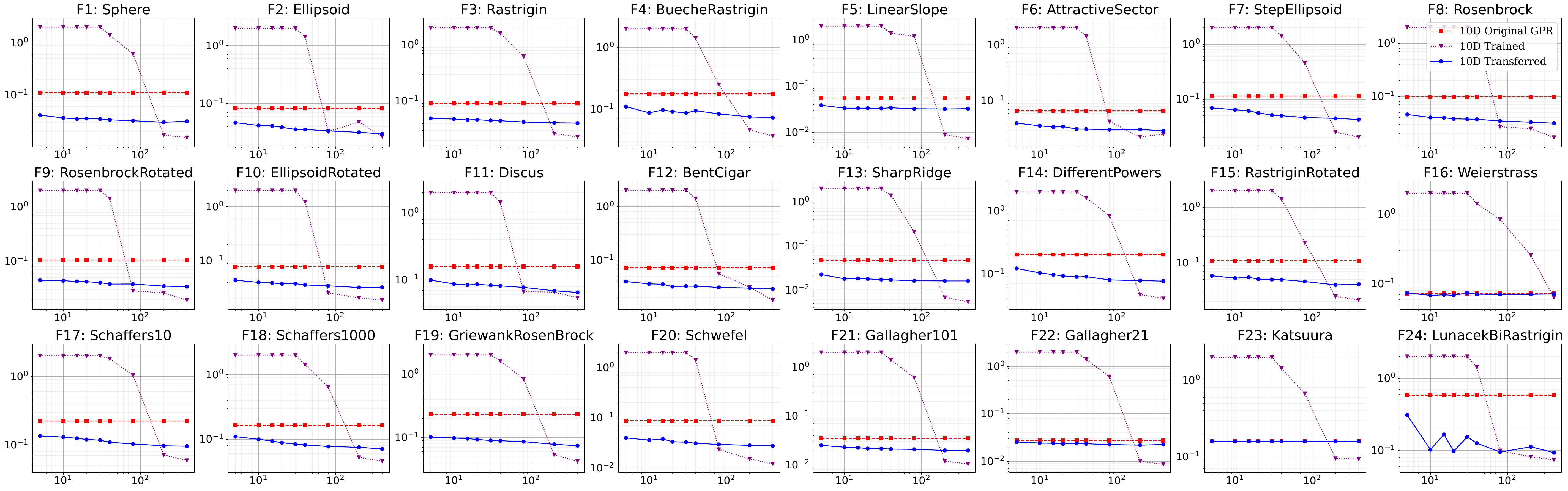}
  \caption{The SMAPE values ($y$-axis) for the original GPR, transferred GPR, and GPR trained solely on the transfer dataset are plotted against the transfer dataset sizes ($x$-axis: 5, 10, 15, 20, 30, 40, 80, 200, 400) for 10D BBOB functions. The analysis combines a beta CDF warping function (approximating a logarithmic transformation) with an affine transformation.}
  \Description{The SMAPE values (displayed on the $y$-axis) are visualized for three different models: the original GPR, the transferred GPR, and the one trained solely on the transfer dataset.}
  \label{figure:GPRSMAPEplot_logarithmic_dim10}
\end{figure*}

\begin{figure*}[!htbp]
  \centering
  \includegraphics[width=\textwidth]{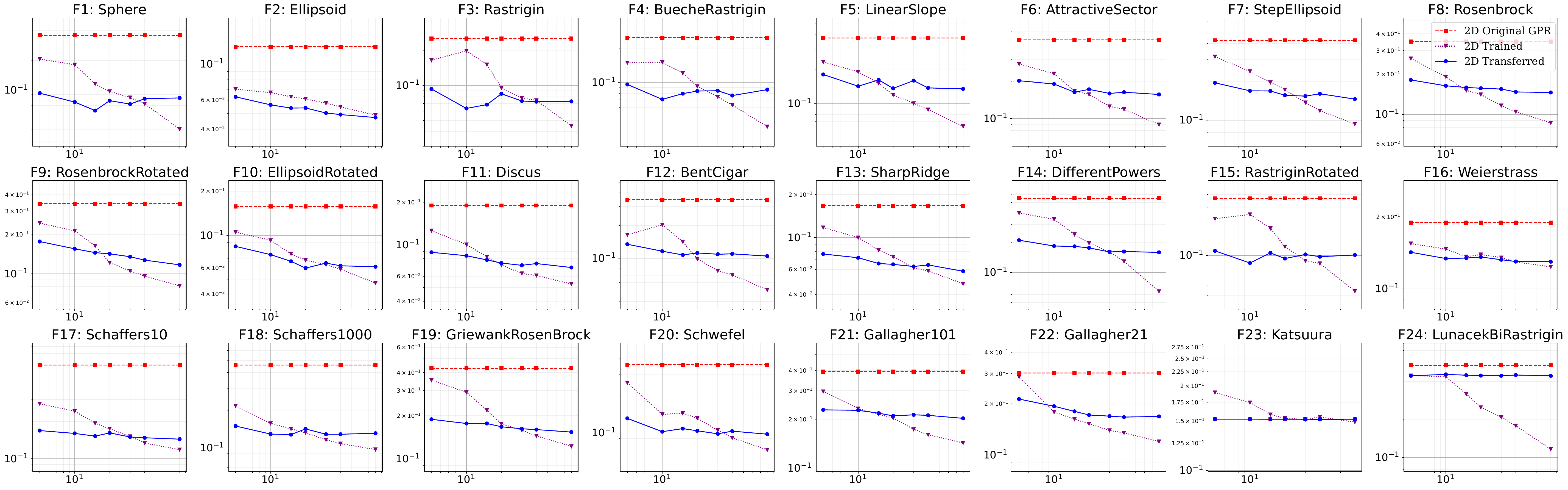}
  \caption{The SMAPE values ($y$-axis) for the original GPR, transferred GPR, and GPR trained solely on the transfer dataset are plotted against the transfer dataset sizes ($x$-axis: 5, 10, 15, 20, 30, 40, 80) for 2D BBOB functions. The analysis combines a beta CDF warping function (approximating a sigmoidal transformation) with an affine transformation.}
  \Description{The SMAPE values (displayed on the $y$-axis) are visualized for three different models: the original GPR, the transferred GPR, and the one trained solely on the transfer dataset.}
  \label{figure:GPRSMAPEplot_sigmoidal_dim2}
\end{figure*}

\begin{figure*}[!htbp]
  \centering
  \includegraphics[width=\textwidth]{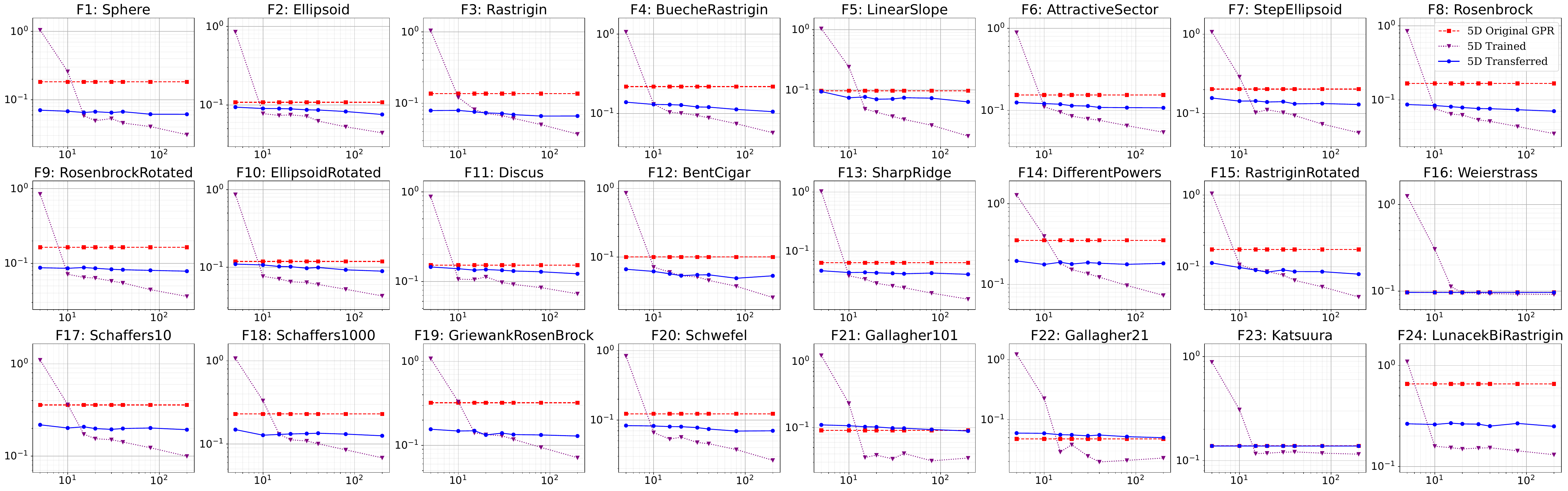}
  \caption{The SMAPE values ($y$-axis) for the original GPR, transferred GPR, and GPR trained solely on the transfer dataset are plotted against the transfer dataset sizes ($x$-axis: 5, 10, 15, 20, 30, 40, 80, 200) for 5D BBOB functions. The analysis combines a beta CDF warping function (approximating a sigmoidal transformation) with an affine transformation.}
  \Description{The SMAPE values (displayed on the $y$-axis) are visualized for three different models: the original GPR, the transferred GPR, and the one trained solely on the transfer dataset.}
  \label{figure:GPRSMAPEplot_sigmoidal_dim5}
\end{figure*}

\begin{figure*}[!htbp]
  \centering
  \includegraphics[width=\textwidth]{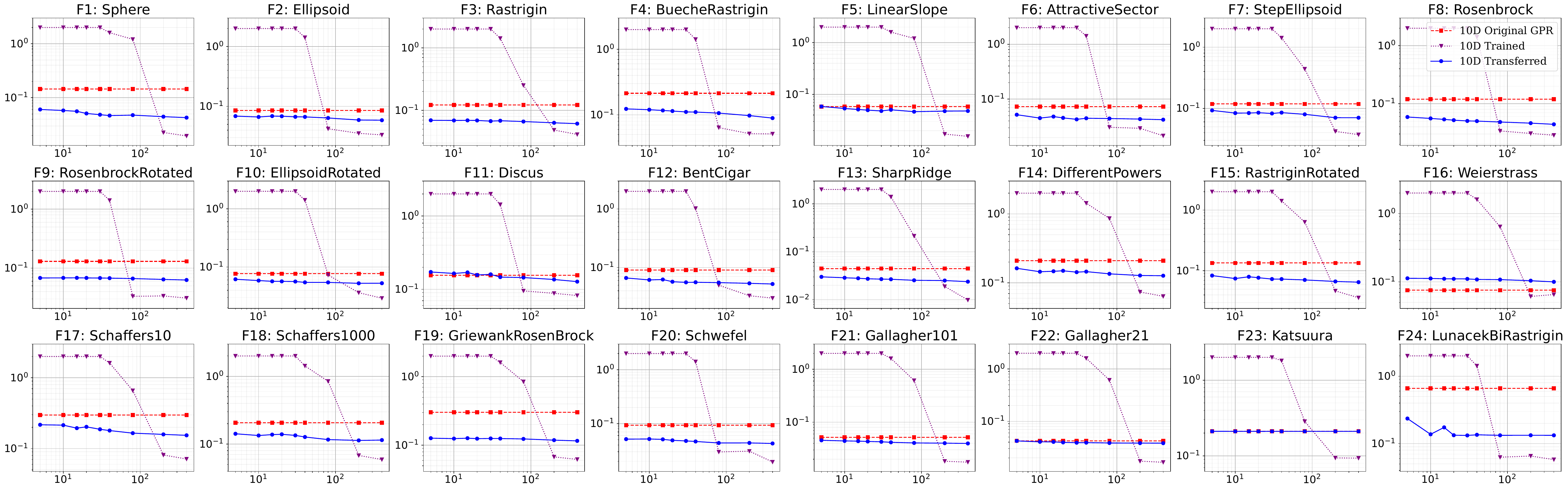}
  \caption{The SMAPE values ($y$-axis) for the original GPR, transferred GPR, and GPR trained solely on the transfer dataset are plotted against the transfer dataset sizes ($x$-axis: 5, 10, 15, 20, 30, 40, 80, 200, 400) for 10D BBOB functions. The analysis combines a beta CDF warping function (approximating a sigmoidal transformation) with an affine transformation.}
  \Description{The SMAPE values (displayed on the $y$-axis) are visualized for three different models: the original GPR, the transferred GPR, and the one trained solely on the transfer dataset.}
  \label{figure:GPRSMAPEplot_sigmoidal_dim10}
\end{figure*}

\begin{figure*}[!htbp]
  \centering
  \includegraphics[width=\textwidth]{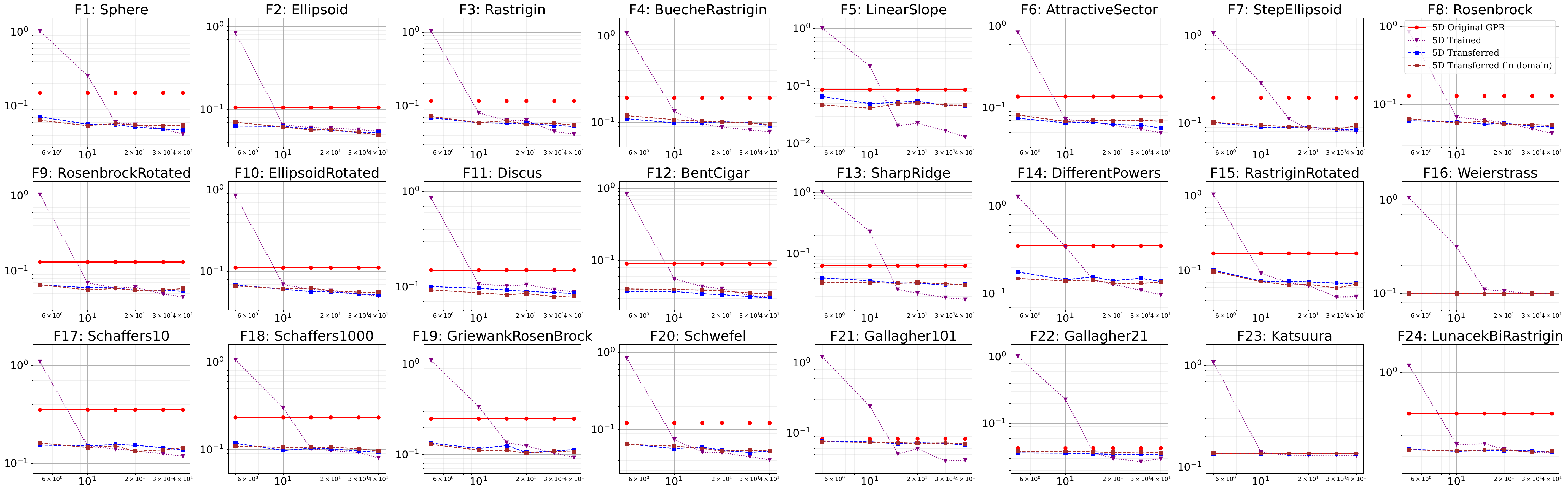}
  \caption{Ablation study presents results for the ``in domain'' scenario, where only transfer data—sampled from the target domain and mapped back into the original domain after transformation—is used for training. SMAPE values ($y$-axis) are shown for the original GPR, transferred GPR, and a model trained solely on the transfer dataset, plotted against transfer dataset sizes ($x$-axis: 5, 10, 15, 20, 30, 40) for 5D BBOB functions. The analysis combines a beta CDF warping function (approximating an exponential transformation) with an affine transformation.}
  \Description{Ablation study. This figure includes results for a scenario where only the transfer data—sampled directly from the target domain and mapped back into the original domain after transformation (referred to as ``in domain'')—is used for training.}
  \label{figure: GPRSMAPEplot_ablation_in_domain_data_dim5}
\end{figure*}
\begin{figure*}[!ht]
  \centering
  \includegraphics[width=0.75\textwidth]{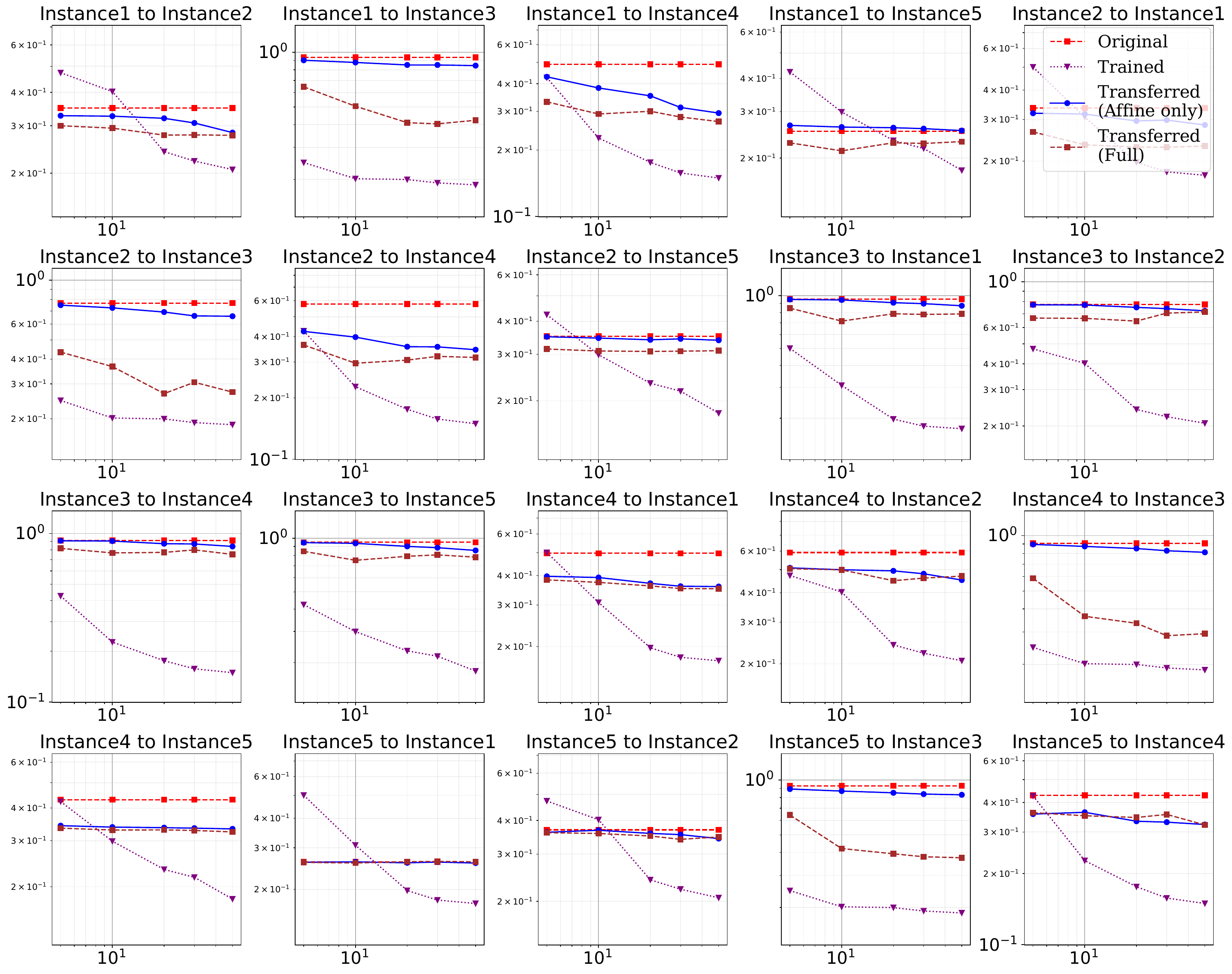}
  \caption{The study evaluates SMAPE values ($y$-axis) for four GPR models on an automotive industry benchmark: the original GPR model, a transferred GPR with an assumed affine transformation (``Transferred (Affine only)''~\cite{PanVerLopBac2024transfer}), a transferred GPR model using the proposed method (``Transferred (Full)''), and a model trained solely on the transfer dataset. SMAPE values are plotted against transfer dataset sizes ($x$-axis: 5, 10, 20, 30, and 50).
  }
  \Description{The study evaluates SMAPE values (displayed on the $y$-axis) for four different Gaussian process regression variants on a real-world benchmark within the automotive industry.}
  \label{figure:GPR_SMAPEplot_real_world_vehicle_2D}
\end{figure*}

\end{document}